\documentclass[11pt]{article}

\usepackage[T1]{fontenc}
\usepackage{lmodern}
\usepackage[margin=1in]{geometry}
\usepackage{amsmath,amssymb}
\usepackage{mathtools}
\usepackage{array,booktabs,tabularx}
\usepackage{adjustbox}
\usepackage{graphicx}
\usepackage{float}
\usepackage{xcolor}
\usepackage{microtype}
\usepackage[numbers,sort&compress]{natbib}
\usepackage{hyperref}
\usepackage{placeins}
\usepackage{public_report}

\newcolumntype{Y}{>{\raggedright\arraybackslash}X}

\title{Format-Aware Fusion for Fast FP4 Pretraining}
\reportsubtitle{Native MXFP4 and NVFP4 Training on FP4-Compatible Hardware}
\author{Robert Hu\\[1mm]
\small\mdseries Correspondence:
\href{mailto:robert.stats.hu@gmail.com}{\nolinkurl{robert.stats.hu@gmail.com}}}
\date{September 2026\\[3mm]
\small \textcopyright{} 2026 Robert Hu. Original manuscript and figures
licensed under \href{https://creativecommons.org/licenses/by/4.0/}{CC BY 4.0}.}
\hypersetup{
  colorlinks=true,
  linkcolor=ReportInk,
  citecolor=ReportAccentDark,
  urlcolor=ReportAccentDark,
  pdftitle={Format-Aware Fusion for Fast FP4 Pretraining},
  pdfauthor={Robert Hu},
  pdfsubject={Native MXFP4 and NVFP4 training with format-aware fusion on FP4-compatible hardware},
  pdfkeywords={FP4 training, MXFP4, NVFP4, format-aware fusion, stochastic rounding, Llama}
}

\newcommand{\mxfp}{MXFP4}
\newcommand{\nvfp}{NVFP4}
\newcommand{\localcta}{CTA-local NVFP4}
\newcommand{\tk}{ThunderKittens}

\begin{document}

\maketitle
\begin{abstract}
Four-bit floating-point (FP4) Tensor Cores accelerate matrix multiplication,
but scale computation, operand packing, layout construction, and saved
backward state can erase the gain. We present \emph{format-aware fusion}, which
co-designs each quantization producer with its scale domain and consumer layout
for native \mxfp{}, global \nvfp{}, and cooperative-thread-array-local
\nvfp{}. We evaluate Llama-3-family 8B pretraining through 160 billion tokens
using bfloat16 output projections and compiled cross entropy. In matched
same-accelerator probes, bfloat16 and Transformer Engine \nvfp{} reach 18.8K
and 27.6K tokens/s/GPU, while our fastest custom route reaches 37.9K. \mxfp{}
with row-gradient stochastic rounding and fixed-sign 32-value Hadamard
weight-gradient preconditioning reaches 37.2K tokens/s/GPU (86.3\% bfloat16
model FLOP utilization) and ends 2.11\% above the raw bfloat16 training-loss
endpoint. A Transformer Engine recipe with four final bfloat16 blocks ends
0.87\% above bfloat16 at 27.1K tokens/s/GPU. Downstream rankings differ from
training-loss rankings, showing that FP4 outcomes depend jointly on scale
contract, operand, and execution path.
\end{abstract}

\section{Introduction}

Four-bit floating-point (FP4) Tensor Cores accelerate general matrix
multiplication (GEMM), but each training layer must also produce scales, pack
values, build forward and backward layouts, and retain autograd state. When
separate kernels move bfloat16 (BF16) intermediates through high-bandwidth
memory (HBM), this surrounding work can erase the GEMM speedup
\cite{nvidia2025nvfp4}.

We address this gap with \emph{format-aware fusion}: each producer completes
the scale shared by its output values while emitting the packed layouts and
saved state needed by the following matrix multiplications. We use
\emph{route} to mean a complete numerical recipe and execution path, not only
an FP4 datatype. \mxfp{} completes a local 32-value scale, global
\nvfp{} waits for a tensor-wide maximum, and cooperative-thread-array-local
(CTA-local) \nvfp{} completes two outer scales inside the tile that owns the
matrix multiplication. The profitable fusion boundary therefore differs by
format.

We evaluate the resulting paths in Llama-3-family 8B pretraining at sequence length
8,192 and global batch 512. Same-accelerator probes measure every route with a
BF16 output projection and compiled cross entropy (CE), while long
trajectories cover up to 160B tokens. The fastest custom route reaches 37.9K
tokens/s/GPU. The \mxfp{} route with row-gradient stochastic rounding (SR)
and fixed-sign 32-value Hadamard (H32) weight-gradient (Wgrad)
preconditioning reaches 37.2K
tokens/s/GPU and ends 2.11\% above the raw BF16 endpoint. The TE-native and
four-final-BF16-block trajectories end 1.34\% and 0.87\% above raw BF16,
respectively, at measured throughputs of 27.6K and 27.1K tokens/s/GPU.

Table~\ref{tab:prior-work-boundary} situates the implementation relative to
established FP4 practice. We contribute:

\begin{itemize}
  \item \emph{Numerical parameterization.} We specify the FP4 contract per
        operand: upward-rounded \mxfp{} scales for activations and gradients;
        an adaptive finer scale for $32\times32$ weight tiles; separate FP32
        row- and column-tile scale factors for \localcta{}; stochastic
        rounding only for the data-gradient multiplication; and paired
        fixed-sign 16- or 32-value Hadamard preconditioning only for the
        weight-gradient multiplication.

  \item \emph{Format-aware fusion and native implementation.} We implement
        custom \mxfp{}, custom global \nvfp{}, and \localcta{} producers that
        finish the scale calculation over the required group of values and emit the layouts and saved
        state required by forward and backward. The MXFP4 and CTA-local weight
        producers quantize once and emit both packed weight layouts in the same
        fused operation. Custom global NVFP4 likewise integrates its shared
        $16\times16$ weight encoding and both packed layouts directly into the
        fused path. Backward stages fuse optional Hadamard preconditioning
        with column quantization. The same framework supports depth- and
        operand-wise hybrids.

  \item \emph{End-to-end empirical evaluation.} We implement and evaluate the
        custom pure and hybrid routes alongside BF16 and TE comparators. The
        hybrid designs include depth-wise format assignment and an operand-wise
        assignment motivated by a data-gradient error analysis using identical
        saved tensors. The evidence
        comprises matched throughput probes, trajectories through 160B
        training tokens, checkpoint-matched validation loss on a fixed
        external data stream, and
        downstream evaluation of the complete recipes. We release the
        training and evaluation code, pinned kernel sources, complete recipe
        ledger, publication data, and figure builders at
        \url{https://github.com/MrHuff/mfu-fp4-training};
        Appendix~\ref{app:code-release} describes its reproducibility contract.
\end{itemize}

\section{FP4 Contracts and Numerical Method}

\subsection{Related work}

Low-precision training builds on mixed-precision accumulation and stochastic
rounding \cite{gupta2015limited,micikevicius2018mixedprecision}. Early 4-bit
systems combined integer weights and activations with specialized gradient
formats \cite{sun2020ultralow}. Chmiel et al.'s logarithmic unbiased
quantization (LUQ) targets unbiased neural-gradient quantization over the full
range: it combines stochastic underflow below the smallest nonzero value,
range selection that avoids upper clipping, and stochastic rounding within a
logarithmic FP4 grid \cite{chmiel2023accurate}. This is a stronger construction
than applying ordinary adjacent-bin SR only inside the representable range.
FP8 and microscaling then established floating-point formats with shared block exponents
\cite{micikevicius2022fp8,rouhani2023microexponents,
rouhani2023microscaling,ocp2023mx}. Their behavior depends on more than bit
width: scale format, block geometry, tensor statistics, and the contraction
being quantized all matter
\cite{yang2025microscaling,hu2025designspace,sun2025scaling,fasoli2026finer}.

Recent FP4 pretraining systems make different choices along those axes. Tseng
et al. apply MXFP4 to the two backward matrix multiplications and combine
stochastic rounding with randomized Hadamard transform (RHT) preconditioning
\cite{tseng2025mxfp4}. FP4 All the Way uses round-to-nearest in Fprop and SR in
the backward and update contractions; in Wgrad, it stochastically rounds both
the neural-gradient and activation operands \cite{chmiel2025fp4alltheway}.
NVIDIA's NVFP4 recipe combines two-dimensional weights, gradient stochastic
rounding, paired Wgrad preconditioning, and selected higher-precision linears
\cite{nvidia2025nvfp4}. Quartet and FP4 All the Way study fully low-precision
paths, while subsequent work develops native
full-pipeline MXFP4 and unbiased NVFP4 gradient estimation
\cite{castro2025quartet,chmiel2025fp4alltheway,cim2026nativemxfp4,
panferov2026quartetii}.
Other FP4 studies examine optimizer and oscillation effects, adaptive block
scales, outlier dynamics, shrinkage bias, and quantization beyond the linear
operators
\cite{wang2025fp4,zhou2025exploringfp4,chen2025tetrajet,
chen2025tetrajetv2,cook2025fouroversix,dong2026outliers,
zhao2026ufp4,ding2026fullstack,agrusa2026whatmatters}. Complementary systems
explore spectral-domain FP4, alternative four-bit formats, and native FP4
training for mixture-of-experts models
\cite{cao2026metis,taghian2026hifloat4,du2026halfs,zhang2026hoppermoe}.
UE5M3 FP4 reallocates the unused sign bit of an E4M3 block scale to a fifth
exponent bit. Hu et al. combine this wider unsigned scale with periodic
sample-and-hold tensor scaling and selective backward SR, while omitting RHT
and using FP4 in every eligible internal linear
\cite{hu2026ue5m3}.

We abbreviate the forward, input-gradient, and weight-gradient matrix
multiplications as Fprop, Dgrad, and Wgrad, respectively.

Three mechanisms directly shape this work. A weight encoding can
remain numerically consistent under transposition
\cite{nvidia2025nvfp4,rahimifar2026transposition,cim2026nativemxfp4}.
Stochastic rounding (SR) can remove within-bin rounding bias, while its placement
and treatment of underflow and saturation---clipping to the largest
representable magnitude---remain part of the quantizer
contract \cite{gupta2015limited,chmiel2023accurate,ozkara2025sr}. Paired Hadamard preconditioning spreads
outliers while preserving the unquantized dot product
\cite{ashkboos2024quarot,tseng2025mxfp4,nvidia2025nvfp4,
cim2026nativemxfp4}. We specialize where these mechanisms act in each
contraction and realize the complete operand contract inside fused training
kernels.

The systems contribution is to optimize the quantizer, its physical layouts,
the consuming matrix multiplication, and retained backward state as one
execution boundary. This follows the broader principle that data movement and
producer--consumer structure determine realized model speed
\cite{dao2022flashattention,dao2023flashattention2,hsu2024liger,
wijmans2024cutcrossentropy,spector2024thunderkittens,guo2026coda}.

E2M1 denotes the four-bit value format with one sign bit, two exponent bits,
and one fraction bit. E8M0 is a power-of-two scale format, E4M3 is an
eight-bit floating-point scale format, and \emph{amax} denotes the maximum
absolute value. A 2D16 or 2D32 weight encoding shares one numerical scale over
a $16\times16$ or $32\times32$ tile and reuses it when the weight is viewed in
either orientation. H16 and H32 denote Hadamard transforms that mix 16 or 32
values. Table~\ref{tab:prior-work-boundary} pairs each existing method with its
implementation in this work; the following subsections explain these choices
in detail.

\begin{table}[tbp]
  \centering
  \footnotesize
  \caption{Existing numerical and systems methods and their parameterization,
  composition, or implementation in this work.}
  \label{tab:prior-work-boundary}
  \begin{tabularx}{\textwidth}{@{}lYY@{}}
    \toprule
    Component & Existing method & Implementation in this work \\
    \midrule
    MXFP4 scale policy
      & E8M0 forces the block scale to a power of two. Rounding the exponent
        up keeps the block maximum within the largest finite E2M1 magnitude,
        6; rounding
        it down halves the decoded grid spacing while clamping the upper tail
        to that endpoint
        \cite{ocp2023mx,cim2026nativemxfp4}
      & Scale direction is explicit and operand-specific. Activations and
        gradients use upward, encode-centric scales; 2D32 weights admit the
        next finer scale only under the fixed tile-maximum test. Fused
        producers implement both rules \\
    CTA-local outer scaling
      & NVFP4 uses one tensor-level FP32 outer scale together with an E4M3
        scale for each 16 values \cite{ocp2023mx,nvidia2025nvfp4}
      & We factor the outer correction into FP32 row- and column-tile factors
        that remain fixed along K. Their product is applied after the full
        Tensor Core accumulation, replacing the tensor-wide maximum-magnitude
        reduction
        with tile-local scale completion \\
    2D weight contract
      & NVFP4 uses orientation-consistent 2D16 weights, and existing MXFP4
        systems describe transpose-consistent 2D32 weights
        \cite{nvidia2025nvfp4,nvidia2026te,rahimifar2026transposition,
        cim2026nativemxfp4}
      & Pure-format and depth-wise routes preserve one weight encoding across
        Fprop and Dgrad. Custom global NVFP4 returns both packed views under one 2D16
        encoding; MXFP4 and CTA-local fused producers quantize once and emit
        both layouts. The operand hybrid instead emits distinct encodings from
        the same BF16 tile \\
    Stochastic-rounding placement
      & Chmiel et al. use adjacent-bin stochastic rounding in Dgrad and on
        both Wgrad operands; NVIDIA applies gradient SR to both Dgrad and Wgrad
        \cite{chmiel2025fp4alltheway,nvidia2025nvfp4}
      & One producer creates two consumer-specific $dY$ representations: a
        stochastically rounded row-facing operand for Dgrad and a deterministic column-facing
        representation for Wgrad \\
    Wgrad preconditioning
      & Existing FP4 recipes use paired randomized or deterministic Hadamard
        transforms
        \cite{tseng2025mxfp4,nvidia2025nvfp4,cim2026nativemxfp4}
      & In signed-Hadamard routes, fixed-sign H16/H32 reuses the fixed
        16-entry sign pattern from the pinned TE implementation; the plain-H16
        hybrid uses no sign flips. Each acts only on paired $X,dY$ Wgrad
        representations and is fused with column quantization. Dgrad and
        the stored weight are unchanged \\
    Format-aware fusion
      & IO-aware fusion and the underlying MXFP4/NVFP4 scale formats are
        documented in earlier systems
        \cite{ocp2023mx,nvidia2025nvfp4,dao2022flashattention,guo2026coda}
      & The fusion boundary is set by when the required scale becomes available.
        Query/key/value, rotary-position, and gated feed-forward stages produce
        packed layouts and saved backward tensors together; custom global NVFP4
        overlaps tensor amax while CTA-local removes that dependency \\
    Operand-wise hybrid
      & Existing recipes vary precision by layer or module
        \cite{zhou2025exploringfp4,nvidia2025nvfp4,
        rahimifar2026transposition}
      & We assign MXFP4 or CTA-local NVFP4 per contraction inside every linear
        and produce both representations together from shared BF16 inputs, motivated
        by comparing Dgrad error on identical saved tensors \\
    \bottomrule
  \end{tabularx}
\end{table}

\subsection{The same four-bit values, three different scale contracts}
\label{sec:scale-contracts}

All three custom routes ultimately present E2M1 operands---one sign bit, two
exponent bits, and one fraction bit---to a Blackwell Tensor Core. Their
numerical and scheduling behavior nevertheless
differs because they assign scales to different domains.  A useful common
notation for one scalar is
\begin{equation}
  \widehat z_i = S\,\delta_{b(i)}
  Q_{\mathrm{E2M1}}\!\left(\frac{z_i}{S\,\delta_{b(i)}}\right),
  \label{eq:hierarchical-quant}
\end{equation}
where $\delta_{b(i)}$ is the effective block dequantization multiplier, $S$
is an optional outer scale, and
$Q_{\mathrm{E2M1}}$ rounds and saturates to E2M1.  The scale choice determines
both quantization error and when the producer may safely release its output.

\begin{table}[htbp]
  \centering
  \caption{The FP4 execution contracts used in this report. CTA-local NVFP4
  changes the outer-scale domain while retaining E2M1 values and E4M3
  block scales.}
  \label{tab:format-contracts}
  \begin{tabularx}{\textwidth}{@{}l c Y Y@{}}
    \toprule
    Route & Block & Scale metadata & Main systems consequence \\
    \midrule
    \mxfp{} & 32 & E8M0 block scale & Local reduction; coarse power-of-two grid \\
    Custom global \nvfp{} & 16 & E4M3 block + FP32 tensor & Requires the completed tensor amax \\
    \localcta{} & 16 & E4M3 block + two FP32 tile-outer scales & Tile-local production; two-sided metadata \\
    \bottomrule
  \end{tabularx}
\end{table}
\FloatBarrier

\subsubsection{MXFP4 encode- and decode-centric scale selection}
\label{sec:mxfp4-scale-direction}

E8M0 represents a power-of-two scale with eight exponent bits and no fraction;
E4M3 is an FP8 scale with four exponent and three fraction bits. In \mxfp{},
$S=1$, $s_b$ is an E8M0 power of two shared by 32 values, and the effective
multiplier in Equation~\ref{eq:hierarchical-quant} is $\delta_b=s_b/6$.
The block maximum therefore produces a scale locally, but adjacent scale
levels differ by exactly two. This leaves a consequential choice about which
neighboring power of two to store.

We call upward exponent rounding \emph{encode-centric} scaling because it
first encloses every source value in a scale that E2M1 can encode without
upper-tail clipping. For a positive, finite block maximum
$a_b=\max_i|z_i|$ within the supported E8M0 exponent range, it computes
\begin{equation}
  s_b=2^{\lceil\log_2 a_b\rceil},\qquad
  q_i=Q_{\mathrm{E2M1}}\!\left(\frac{6z_i}{s_b}\right),\qquad
  \widehat z_i=\frac{s_b}{6}q_i.
  \label{eq:mxfp4-encode-scale}
\end{equation}
The largest magnitude then maps to at most the finite E2M1 endpoint, 6. We
call downward exponent rounding \emph{decode-centric} scaling: it chooses
$s_b=2^{\lfloor\log_2 a_b\rfloor}$ so the decoded multiplier $s_b/6$ is half
as large whenever $a_b$ is not already a power of two. That halves the grid
spacing for values that remain in range, but values above the endpoint
saturate to 6 and lose their excess magnitude. In plain terms,
encode-centric scaling protects the largest value; decode-centric scaling
represents smaller values more finely by accepting clipping at the top of the
block. For example, if $a_b=10$, encode-centric scaling stores $s_b=16$ and
does not clip; decode-centric scaling stores $s_b=8$, halves the spacing, and
clips values whose magnitude exceeds 8.

We expose this direction as part of the MXFP4 route rather than treating it as
an incidental implementation detail. Development-only MXFP4 pretraining
screens favored encode-centric over decode-centric scaling for both activation
and gradient operands. These screens selected the recipe; they are not a
reported quantitative comparison. Accordingly, every reported \mxfp{}
experiment rounds activation and gradient block scales upward, as shown in
Equation~\ref{eq:mxfp4-encode-scale}. This choice protects the largest value
in each block from clipping. The weight rule described next is the only place
where our \mxfp{} recipe sometimes selects the finer downward scale.

\begin{figure}[tbp]
  \centering
  \includegraphics[width=0.96\textwidth]{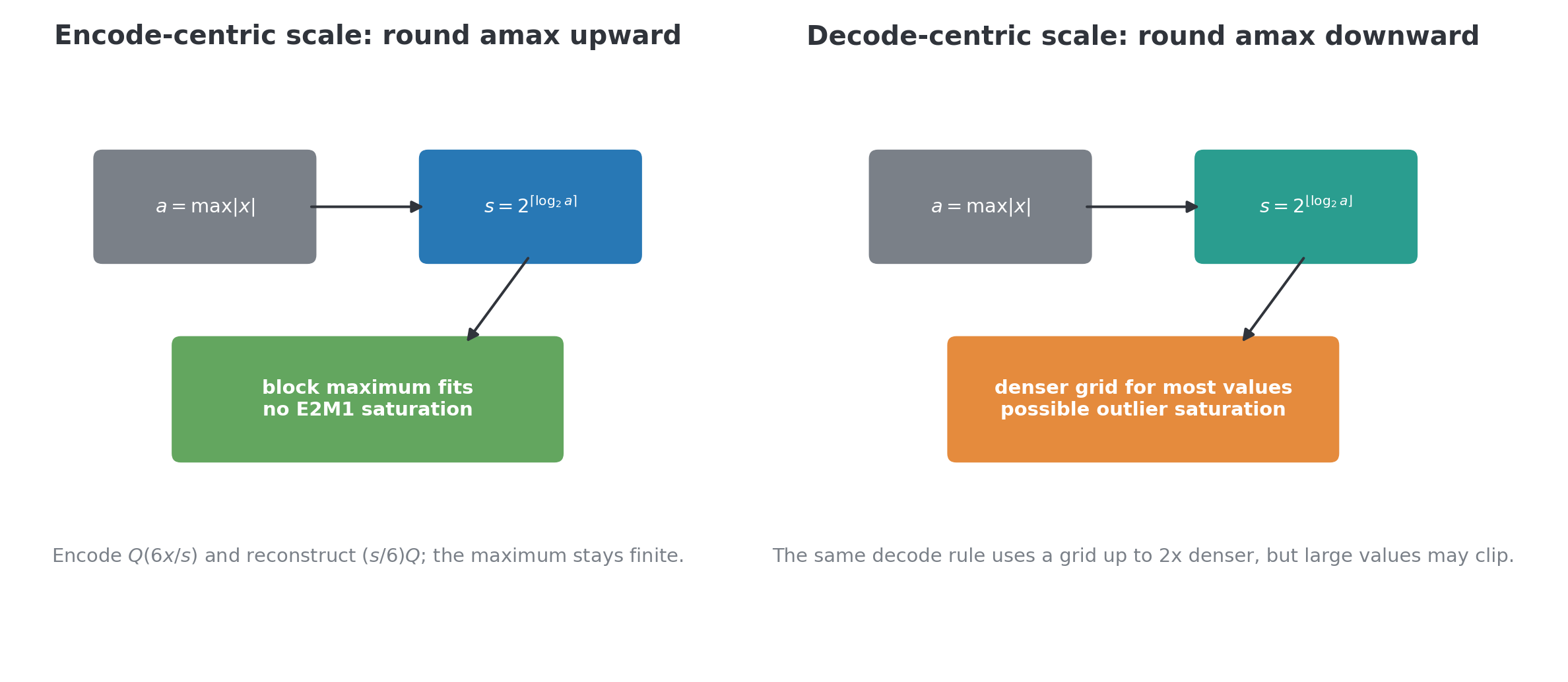}
  \caption{Operand-specific E8M0 scale selection. Encode-centric scaling
  rounds the block maximum upward and keeps every input within the E2M1
  endpoint. For a non-power-of-two maximum, decode-centric scaling rounds
  downward, halves the decoded grid spacing, and clips the block tail. Our
  route applies this range--resolution choice per operand:
  activations and gradients use the encode-centric rule, while 2D32 weights
  select the finer scale only under the fixed $a_T/s_T<0.87$ condition.}
  \label{fig:mxfp4-scale-rounding}
\end{figure}

Weights use a different trade-off. The orientation-consistent 2D32 producer
starts from the safe upward-rounded scale $s_T$ for the maximum $a_T$ over a
$32\!\times\!32$ tile. For a nonzero tile with a representable safe exponent,
it selects $s_T/2$ only when $a_T/s_T<0.87$. The smaller scale halves the
quantization spacing and accepts a fixed, bounded amount of upper-tail
clipping. The same fused producer makes that decision once and emits both
physical consumer layouts. Our MXFP4 implementation makes the
operand-dependent scale policy explicit and performs the scale decision and
dual-layout production inside the same producer.
Figure~\ref{fig:mxfp4-scale-rounding} summarizes the resulting
operand-dependent range--resolution trade-off.

\FloatBarrier

\subsubsection{Global and CTA-local NVFP4 outer scales}

Global \nvfp{} instead uses an E4M3 scale for
each 16 values and an FP32 tensor scale $S$ \cite{ocp2023mx,nvidia2025nvfp4}.
Its local grid is finer, but every block depends on the completed tensor amax,
the maximum absolute value.
Our \emph{custom global NVFP4} path preserves the tensor-global numerical
contract used by TE-native NVFP4 while replacing generic
quantize--layout--GEMM boundaries with custom producers and consumers.

This tensor-wide maximum also creates a scheduling dependency. A GPU divides
the tensor among many cooperative thread arrays (CTAs), which are thread
blocks that may run at different times. Each CTA can find a partial maximum,
but the final tensor maximum is unavailable until all partial results have
been combined. Packing the quantized tensor and starting the dependent matrix
multiplication must wait for that result. The wait forms a tensor-wide
synchronization boundary within the GPU and can leave a latency bubble even
when individual tiles are otherwise ready.

\begin{figure}[tbp]
  \centering
  \includegraphics[width=0.98\textwidth]{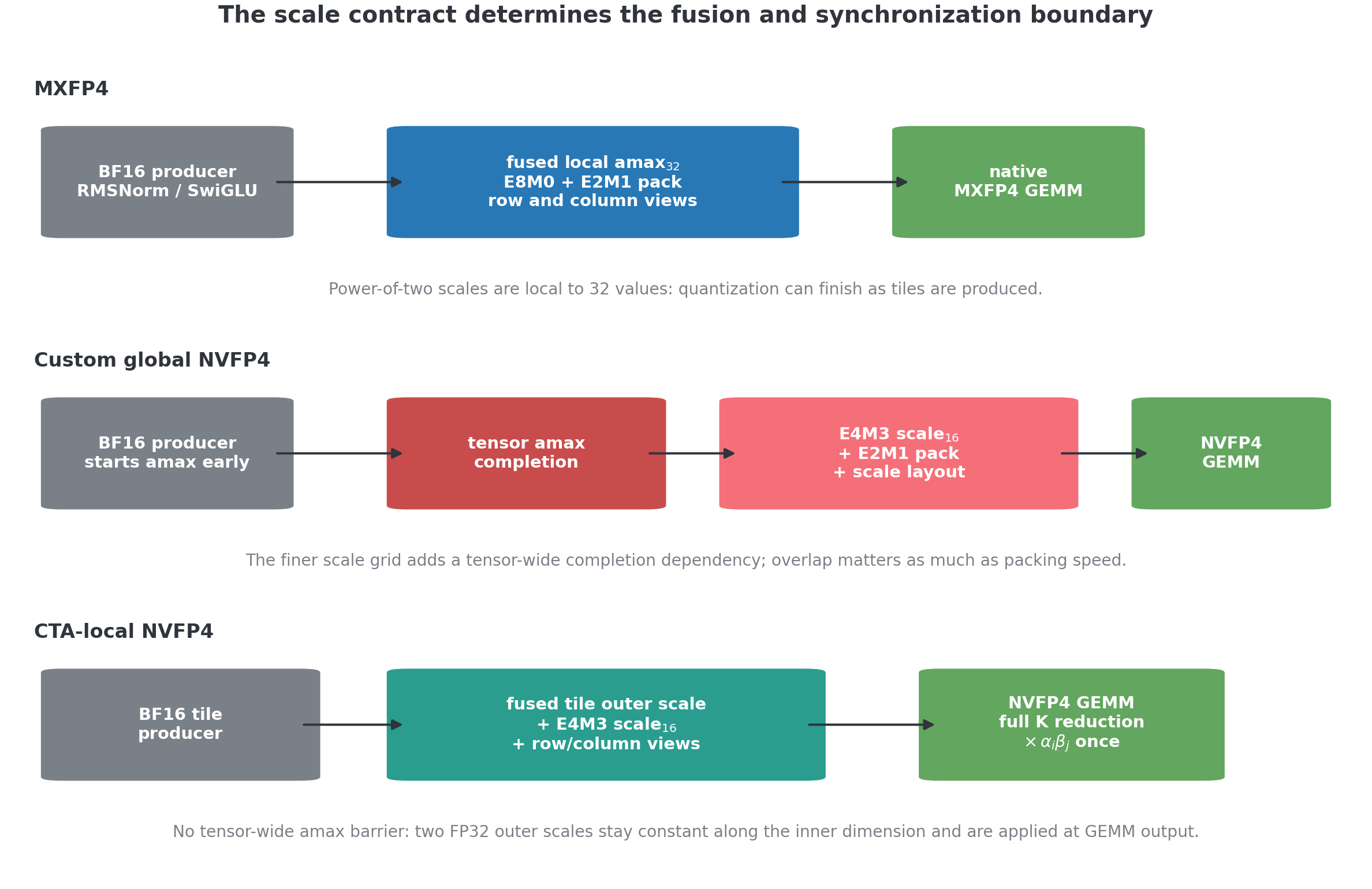}
  \caption{Scale-completion boundaries for the three custom routes. MXFP4
  finalizes a local 32-value scale; global NVFP4 waits for tensor amax;
  CTA-local NVFP4 computes tile-owned row and column outer factors without
  waiting for a tensor-wide reduction. The completion point determines which
  preparation work can be fused with its matrix multiplication.}
  \label{fig:format-execution-routes}
\end{figure}

\begin{figure}[tbp]
  \centering
  \includegraphics[width=0.94\textwidth]{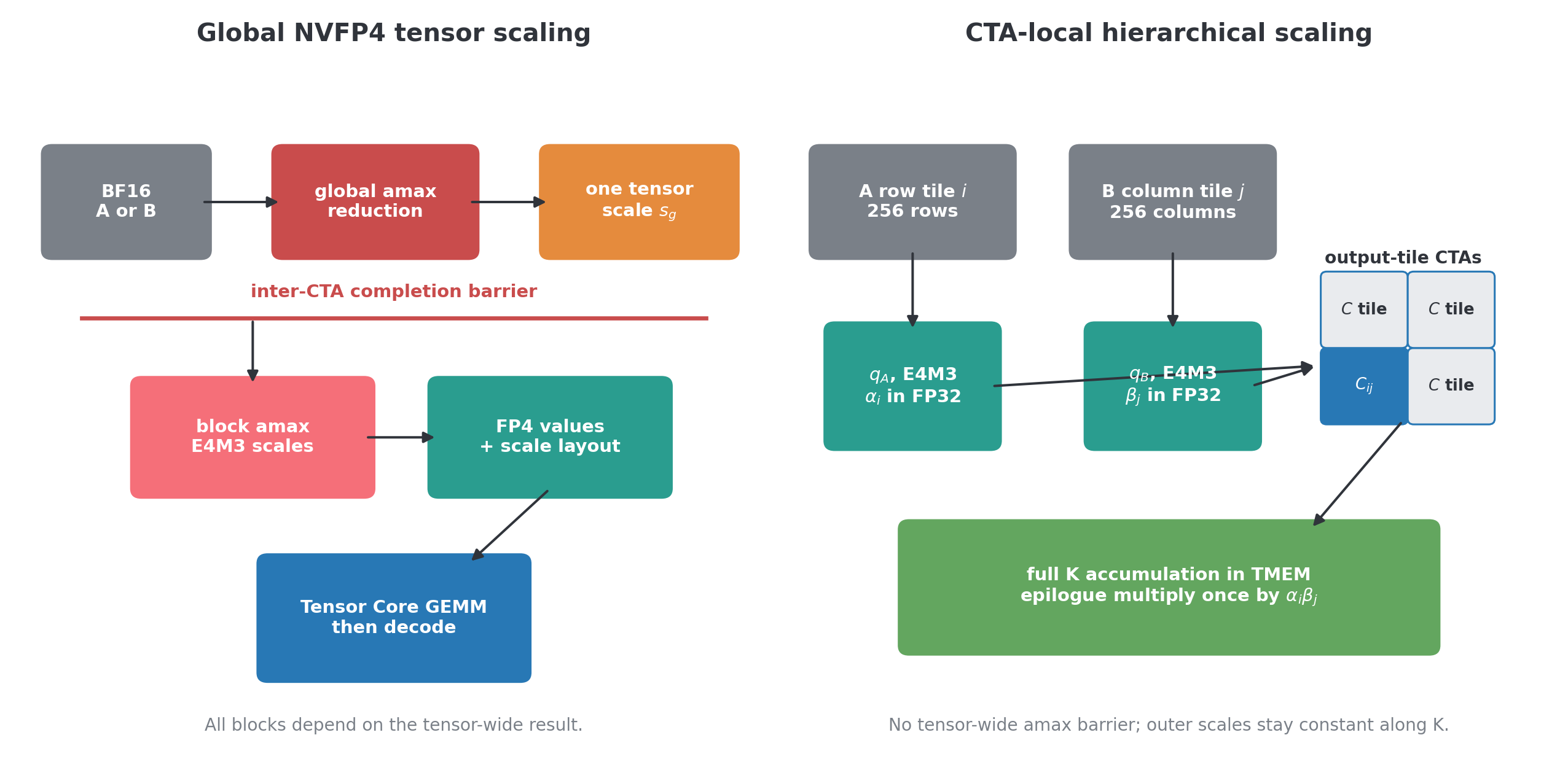}
  \caption{Tensor-global and CTA-local NVFP4 outer-scale domains. Global
  NVFP4 gates block packing on tensor amax. Our CTA-local route carries
  tile-owned row and column outer factors through the full K accumulation and
  applies $\alpha_i\beta_j$ once in the epilogue.}
  \label{fig:localcta-scaling}
\end{figure}

The CTA-local route removes that cross-CTA dependency by changing only the
outer-scale domain. Each CTA computes maxima for the row and column tiles it
already owns, derives their outer factors using synchronization within that
CTA, and can then proceed to its matrix-multiplication tile without waiting
for unrelated CTAs. It does not remove synchronization altogether; it confines
scale completion to one CTA. For one matrix-multiplication output tile, an
A-row tile owns factor $\alpha_i$ and a B-column tile owns factor $\beta_j$.
Both remain fixed over the complete K reduction.  If $\widetilde A$ and
$\widetilde B$ denote E2M1 values decoded with their E4M3 per-block scales,
often called microscales, then
\begin{equation}
  C_{ij}=\alpha_i\beta_j
  \sum_k \widetilde A_{ik}\widetilde B_{kj}.
  \label{eq:localcta}
\end{equation}
The Tensor Core accumulates the inner sum and the epilogue applies
$\alpha_i\beta_j$ once.  This factorization is valid only because the two
outer scales are invariant in $K$, the contraction dimension. Keeping both corrections in FP32 metadata
also avoids quantizing those corrections onto the coarser FP8 scale grid.
Figures~\ref{fig:format-execution-routes} and~\ref{fig:localcta-scaling}
show, respectively, where each format completes its scale state and how the
CTA-local factors replace the tensor-wide outer-scale dependency.

\FloatBarrier

\subsection{Why forward, data gradient, and weight gradient need different views}

\begin{figure}[H]
  \centering
  \includegraphics[width=0.97\textwidth]{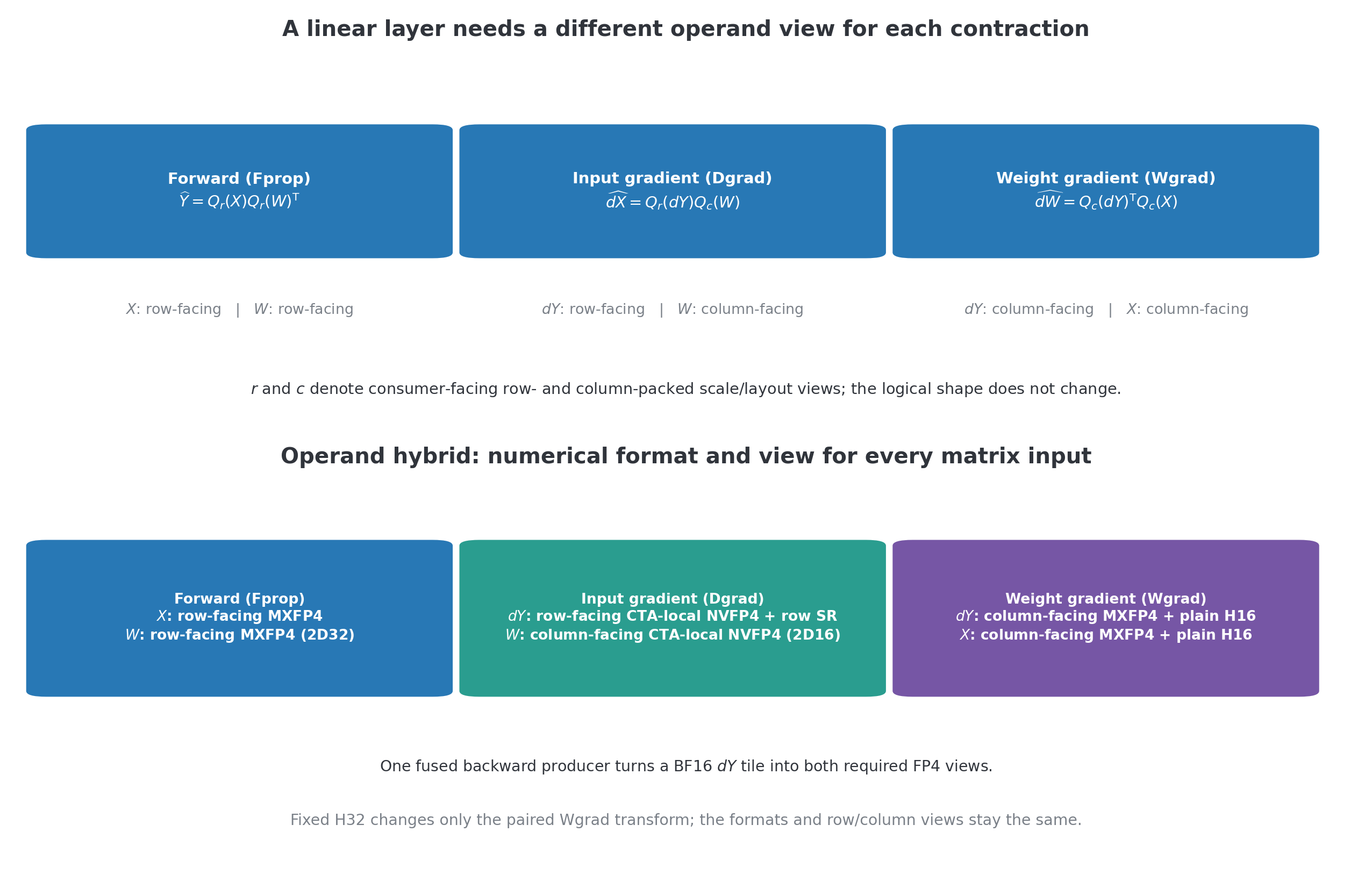}
  \caption{Formats and packed views for the plain-H16 operand hybrid. Here
  $r$ and $c$ denote the consumer-facing row- and column-packed scale/layout
  representations; they do not change a tensor's logical shape. Top: Fprop
  consumes row-facing $X$ and $W$, Dgrad consumes row-facing $dY$ and
  column-facing $W$, and Wgrad consumes column-facing $dY$ and $X$. Bottom:
  Fprop uses MXFP4 $X_r$ and 2D32 $W_r$; Dgrad uses CTA-local NVFP4
  $dY_r$ with row stochastic rounding (SR) and 2D16 $W_c$; Wgrad uses MXFP4
  $dY_c$ and $X_c$ with paired plain-H16 preconditioning. One fused backward
  producer emits both $dY$ representations from the BF16 tile. Wgrad does not
  read $W$ because it produces $dW$. Plain H16 has no sign flips; fixed H32
  changes only the paired Wgrad preconditioner.}
  \label{fig:linear-operand-map}
\end{figure}

For a linear layer $Y=XW^{\mathsf T}$, forward propagation (Fprop), the input
or data gradient (Dgrad), and the weight gradient (Wgrad) consume different
views of the same tensors. Let $Q_r$ denote a row-facing representation and
$Q_c$ a column-facing representation. Both preserve the logical tensor shape;
the subscript identifies the physical layout and scale orientation used by the
consumer. Suppressing decode and scale metadata, training evaluates three
contractions:
\begin{align}
  \widehat Y  &= Q_r(X)Q_r(W)^{\mathsf T},
    &&\text{Fprop}, \label{eq:fprop}\\
  \widehat{dX} &= Q_r(dY)Q_c(W),
    &&\text{Dgrad}, \label{eq:dgrad}\\
  \widehat{dW} &= Q_c(dY)^{\mathsf T}Q_c(X),
    &&\text{Wgrad}. \label{eq:wgrad}
\end{align}
Thus one saved gradient needs a row-oriented representation for Dgrad and a
column-oriented representation for Wgrad; weights likewise participate in two
orientations. ``Quantize the tensor to FP4'' is incomplete unless it also
states which contraction consumes it.
The upper panel of Figure~\ref{fig:linear-operand-map} maps these row- and
column-facing representations to the three contractions.

\FloatBarrier

\subsection{Two-dimensional weights: numerical consistency versus implementation}

\begin{figure}[H]
  \centering
  \includegraphics[width=0.96\textwidth]{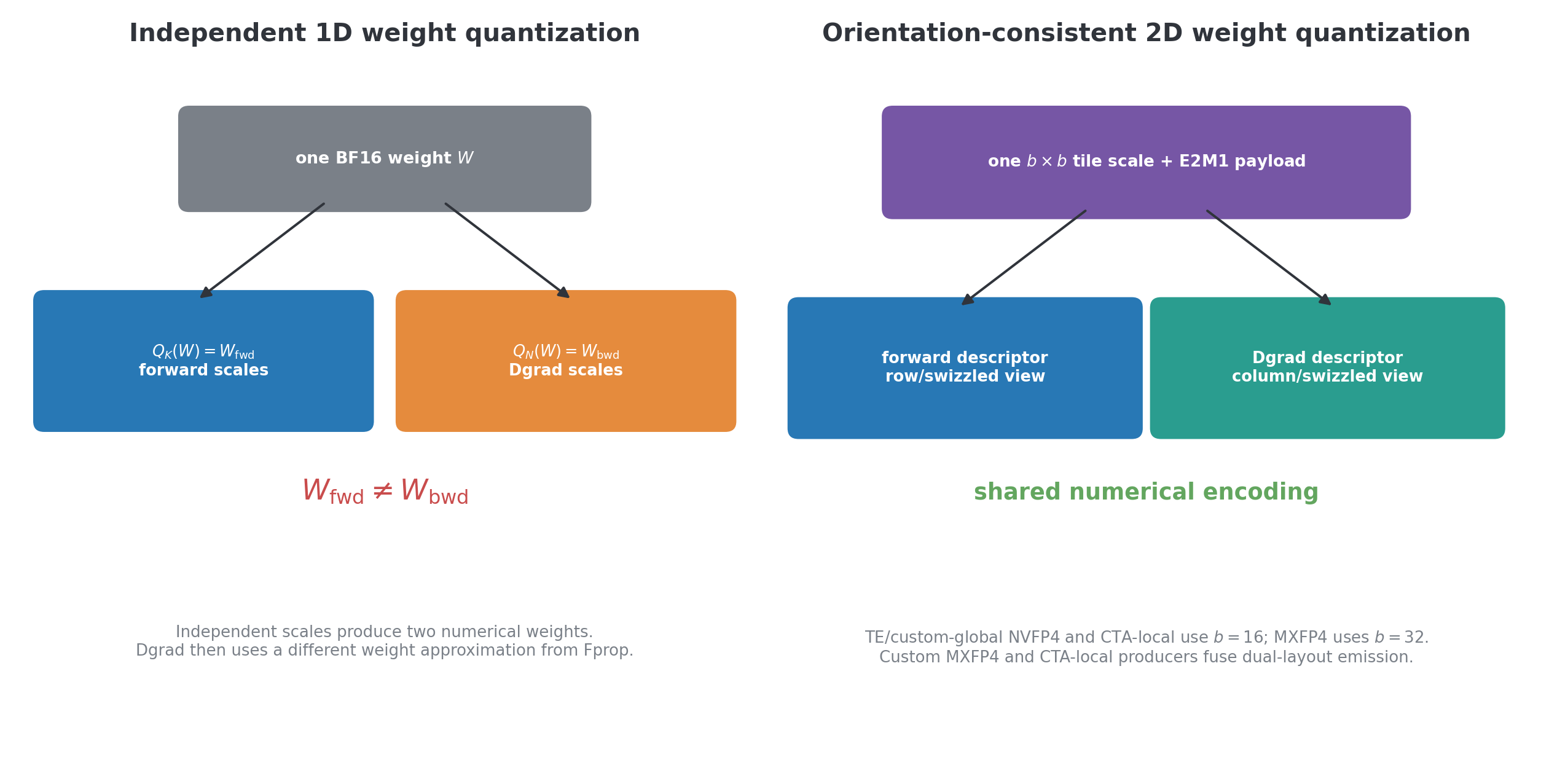}
  \caption{Independent one-dimensional scales can make Fprop and Dgrad consume
  different numerical weights. Orientation-consistent 2D quantization computes
  one tile encoding and exposes it
  through row- and column-oriented memory layouts. Our implementation
  optimization fuses production of those layouts into the custom MXFP4
  and CTA-local producers under their respective scale policies.}
  \label{fig:2d-weight-contract}
\end{figure}

Independent one-dimensional weight quantizers can produce
$W_{\mathrm{fwd}}=Q_K(W)$ for Equation~\ref{eq:fprop} and a different
$W_{\mathrm{bwd}}=Q_N(W)$ for Equation~\ref{eq:dgrad}. Forward and backward
then consume different numerical approximations of the same BF16 weight.
Two-dimensional weight quantization instead assigns one scale to each
$b\times b$ weight tile. Let $q_{\max}$ be the largest finite E2M1 magnitude
and let $\operatorname{scale}_{\cal S}$ denote the route's scale-selection
rule on its representable scale set $\cal S$. Then
\begin{equation}
  a_{pq}=\max_{i\in p,j\in q}|W_{ij}|, \qquad
  s_{pq}=\operatorname{scale}_{\cal S}
  \!\left(\frac{a_{pq}}{S q_{\max}}\right), \qquad
  \widehat W_{ij}=S\,s_{pq}
  Q_{\mathrm{E2M1}}\!\left(\frac{W_{ij}}{S\,s_{pq}}\right),
  \label{eq:2d-weight}
\end{equation}
and reuses that numerical encoding in both orientations. The row and column
layouts can differ in memory, but their decoded weight values agree. This is
the consistency motivation identified by NVIDIA and by work on
transposition-invariant block quantization
\cite{nvidia2025nvfp4,rahimifar2026transposition}.
Equivalently, the numerical invariant is
\begin{equation}
  Q_r^{\mathrm{2D}}(W)\equiv Q_c^{\mathrm{2D}}(W)\equiv\widetilde W,
  \label{eq:2d-weight-invariant}
\end{equation}
even though the two memory layouts contain differently ordered bytes.
Figure~\ref{fig:2d-weight-contract} contrasts independent one-dimensional
quantization with this shared numerical encoding.

Transformer Engine (TE), custom global NVFP4, and the CTA-local path all preserve one
numerical weight encoding across the two $16\times16$ consumer orientations.
In TE's 2D16 contract, one scale belongs to each $16\times16$ weight tile and
is replicated into the $1\times16$ scale layouts required by the Tensor
Core consumers \cite{nvidia2025nvfp4,nvidia2026te}. ``2D'' therefore describes
shared numerical scale ownership across orientations; the matrix-multiplication
shape is unchanged. TE and
custom global NVFP4 use the tensor-level NVFP4 outer scale; the custom route emits the
packed values, block scales, and both physical weight layouts directly.
CTA-local NVFP4 retains the 2D16 consistency invariant. Its weight producer
finds the local weight maxima and E4M3 block scales, folds the relative
corrections into those scales, and publishes one common FP32 weight outer
factor for both orientations. Activations and gradients continue to use the
tile-owned row and column factors in Equation~\ref{eq:localcta}. Our \mxfp{} path applies the same
orientation-consistency objective to $32\times32$ MX tiles. Tseng et al.'s
backward-only MXFP4 method instead
quantizes each operand along the reduction axis of the current contraction
\cite{tseng2025mxfp4}. Transpose-consistent 2D32 MXFP4 is also described in
existing systems \cite{rahimifar2026transposition,cim2026nativemxfp4}.

The numerical 2D contract follows the cited 2D16 and 2D32 methods. Our
implementation optimization is producer fusion: the MXFP4 and CTA-local
producers each load the BF16 weight tile once,
complete the route's scale reduction, quantize one numerical encoding, and
emit the row- and column-oriented physical layouts within the same producer
boundary. The MXFP4 producer incorporates
the adaptive 2D32 scale rule in Section~\ref{sec:mxfp4-scale-direction}; the
CTA-local producer combines its 2D16 encoding with local maximum and scale
calculation, then normalizes both layouts to a common weight outer factor. This construction
removes a second independent scale decision and guarantees that Fprop and
Dgrad decode the same numerical weight even though their physical layouts
differ.

Equation~\ref{eq:2d-weight-invariant} applies to the pure routes and the
depth-wise hybrid. The operand-wise hybrid is an explicit exception: it reads
the BF16 weight once, then emits a 2D32 MX weight encoding for Fprop and a 2D16
CTA-local weight encoding for Dgrad. Wgrad does not consume $W$; it separately
quantizes $X$ and $dY$ with MXFP4. This gives each contraction a different
error profile, so end-to-end training must evaluate the tradeoff.

\subsection{Stochastic rounding and signed Hadamard preconditioning}

For adjacent representable values $q_-$ and $q_+$, data stochastic rounding
(SR) uses
\begin{equation}
  \Pr[Q_{\mathrm{SR}}(z)=q_+]
  =\frac{z-q_-}{q_+-q_-}, \qquad
  \Pr[Q_{\mathrm{SR}}(z)=q_-]=1-\Pr[Q_{\mathrm{SR}}(z)=q_+],
  \label{eq:sr}
\end{equation}
which is unbiased between the two grid points before saturation
\cite{gupta2015limited,ozkara2025sr}. This is the ideal probability rule;
implementations that use too few random bits can introduce bias
\cite{fitzgibbon2025fewbits}. Equation~\ref{eq:sr} is ordinary adjacent-bin SR,
not LUQ. LUQ also stochastically handles values below the logarithmic FP
minimum and selects the range to avoid clipping \cite{chmiel2023accurate}. Our
microscaled E2M1 paths can saturate, so we claim removal of deterministic bias
between adjacent finite values, not unbiasedness of the complete saturated
quantizer.

The placement also differs. FP4 All the Way applies SR to the neural gradient
entering Dgrad and to both the neural-gradient and activation operands entering
Wgrad \cite{chmiel2025fp4alltheway}. NVIDIA recommends round-to-nearest for
weights and activations and SR for the gradient input to both Dgrad and Wgrad
\cite{nvidia2025nvfp4}. Our custom long routes are narrower: data SR is enabled
only on the row-oriented $dY$ representation consumed by Dgrad. The Wgrad
column representation of $dY$, weights, activations, and all scale values
remain deterministic. This axis-selective placement defines the custom
backward operand contract.
The row/column separation is visible in
Figure~\ref{fig:linear-operand-map}; Figure~\ref{fig:qkv-ffn-fusion} locates
the stochastically rounded Dgrad input and paired transformed Wgrad inputs in
the fused backward schedule.

Hadamard preconditioning addresses a different failure mode. One outlier can
set the scale for a whole block, leaving fewer useful E2M1 values for the other
entries. A Hadamard transform mixes the block with additions and subtractions
before quantization, spreading a large coordinate over the block. In Wgrad,
this can keep smaller contributions from underflowing to zero. The transform
must be applied to both inputs of the weight-gradient dot product so that it
cancels in exact arithmetic
\cite{ashkboos2024quarot,tseng2025mxfp4,nvidia2025nvfp4}. If $H$ is the
normalized block Hadamard matrix and $R=HD$ uses a sign diagonal $D$, then
\begin{equation}
  dW=dY^{\mathsf T}X=(R dY)^{\mathsf T}(R X),
  \qquad R^{\mathsf T}R=I.
  \label{eq:paired-rht}
\end{equation}
Both Wgrad operands must therefore be transformed as a pair. Transforming only
$dY$ changes the gradient rather than only changing its quantization basis.

The Hadamard matrix itself is deterministic. In an RHT, randomness comes from
the diagonal sign vector $D$, and that vector need not be resampled at every
step. Tseng et al. sample signs and transform both MXFP4 backward contractions,
$(dY,W)$ for Dgrad and $(dY,X)$ for Wgrad, using a 64-value transform in their
main experiments \cite{tseng2025mxfp4}. NVIDIA instead transforms Wgrad only,
using H16 for NVFP4 and H32 in its MXFP4 comparison. Its recipe specifies one
sign vector shared across linear layers throughout training; the TE
implementation used in our experiments stores a fixed 16-entry vector and caches the resulting
transform \cite{nvidia2025nvfp4,nvidia2026te}. H16 and H32 name the number of
values mixed by the transform.

Our kernels reuse that TE 16-entry sign vector as an explicit route
constant. H16 applies the sign pattern once; H32 repeats it across two 16-value halves
before the cross-half butterfly. No signs are drawn at setup or during
training. We therefore call these paths \emph{fixed-sign H16/H32
preconditioning}, rather than randomized Hadamard preconditioning. The
additions and subtractions mix the values; the fixed signs choose which values
are negated first. Applying the same transform to $X$ and $dY$ preserves the
unquantized Wgrad dot product, and the stored weight is never transformed. A
fixed $D$ remains orthogonal and repeatable across layers, steps, and resumes.
It does not inherit a probabilistic concentration guarantee that assumes $D$
is sampled independently of its input. We neither claim that guarantee nor
compare alternative sign vectors.

As in NVIDIA's recipe, Hadamard preconditioning remains Wgrad-only. Our
backward producer creates the two consumer-specific representations in one
boundary: row-facing $dY$ uses SR for Dgrad, while deterministic
column-facing $dY$ is paired with transformed $X$ for Wgrad. In the signed
routes, the fixed-sign transform is performed in registers as part of column
quantization rather than as a separate transform and repack pass. Dgrad and
the stored weight are not transformed. H16 matches the 16-value NVFP4 scale
block, while H32 matches the 32-value MXFP4 scale block
\cite{cim2026nativemxfp4}. The literal sign pattern and its half-block construction are recorded in
Appendix~\ref{app:implementation}. Table~\ref{tab:recipe-state-ledger} states
the sign and rounding policy for every route.

\subsection{Reference NVFP4 recipe versus the routes evaluated here}

The NVIDIA recipe is a bundle, not only the NVFP4 datatype. It recommends
2D16 weights, SR on gradient inputs to Dgrad and Wgrad, a paired H16 signed
Hadamard transform (described as RHT) for
Wgrad, and retaining roughly 15\% of empirically sensitive linears in higher
precision, with most selected linears near the end of the network
\cite{nvidia2025nvfp4}. Three distinctions matter in this report. First, our
TE-native comparator enables TE's 2D, gradient-SR, and signed-Hadamard
mechanisms but places all 32 body blocks in NVFP4; only the common output head
is BF16. Second, the TE comparator with four final BF16 blocks is a separate
NVIDIA-inspired route. It is not a reproduction of NVIDIA's per-linear
sensitivity list. Third, the custom global NVFP4, \mxfp{}, and \localcta{} long routes
use row-Dgrad SR, and only explicitly named routes apply Hadamard
preconditioning. Tables~\ref{tab:recipe-ledger}
and~\ref{tab:recipe-state-ledger} state the complete contract for each result.

\section{Fused Execution}
\label{sec:method}

\subsection{Fuse the complete producer--consumer boundary}

Quantization helps only when its preparation work overlaps the matrix
multiplication rather than extending the critical path. We model the latency
of one FP4 boundary as
\begin{equation}
  T_{\mathrm{boundary}} \approx T_{\mathrm{serial}} +
  \max\!\left(T_{\mathrm{GEMM}},
  T_{\mathrm{reduce}}+T_{\mathrm{quant}}+T_{\mathrm{layout}}\right),
  \label{eq:boundary-cost}
\end{equation}
Here $T_{\mathrm{serial}}$ is work that cannot overlap because of a data
dependency or resource contention. Fusion removes an HBM round trip only when
the producer emits every scale, layout, and saved value required by the next
operation. Accelerating only the GEMM epilogue can leave a global reduction, a
second tensor orientation, or backward recomputation on the critical path.
We implement the fused tensor operations in CUDA and \tk{}
\cite{spector2024thunderkittens}. Following CODA's GEMM-plus-epilogue framing
\cite{guo2026coda}, an operation joins a boundary only when the next stage can
consume its output directly. A Llama block exposes four such boundaries:
root-mean-square normalization (RMSNorm) into the query/key/value (QKV)
projection, QKV into attention, the gated feed-forward network (FFN), and the
residual update into the next RMSNorm. Each has a corresponding backward
boundary.

Our FP4 boundary contract makes the fusion boundary explicit: a producer
completes the group of values that share a scale, emits
both physical orientations needed by Fprop, Dgrad, and Wgrad, and retains the
BF16 and random state required by autograd. Scale completion, layout
production, and saved-state requirements differ across MXFP4, custom global
NVFP4, and CTA-local NVFP4, so one generic quantize--GEMM fusion schedule
cannot represent all three paths.
Figure~\ref{fig:qkv-ffn-fusion} contrasts the serial boundaries with this
format-aware production schedule.

\subsection{What is fused in a Llama block}

\begin{figure}[tbp]
  \centering
  \includegraphics[width=0.99\textwidth]{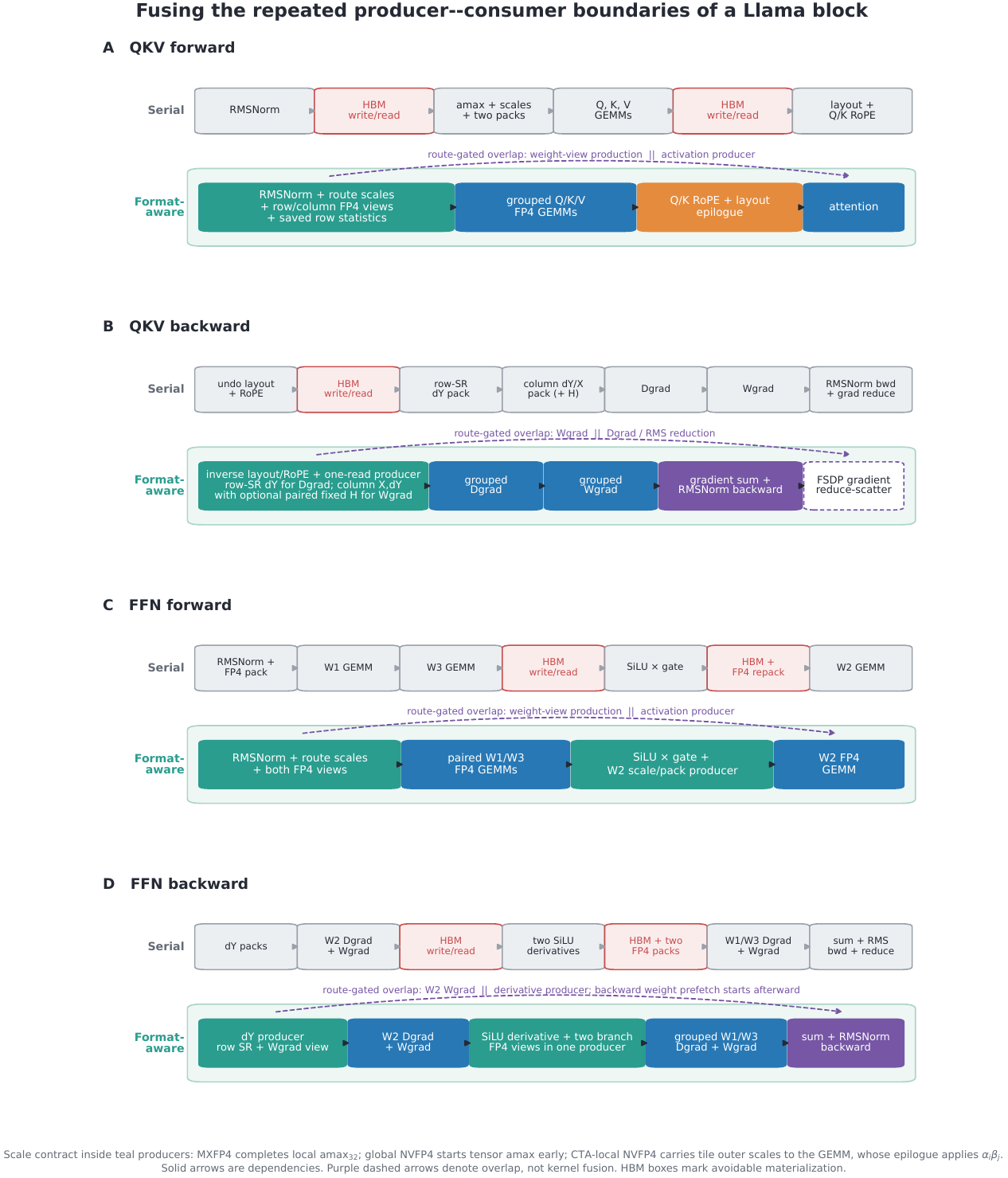}
  \caption{Serial and format-aware execution of the QKV and gated-FFN
  boundaries. Solid boxes and arrows show values produced and consumed within
  one execution boundary; they do not imply one monolithic kernel. Purple
  dashed arrows show stream overlap. Distributed collectives remain separate
  operations even when their launch overlaps local work. In panels B and D,
  row stochastic rounding (SR) applies only to the gradient view consumed by
  Dgrad; a paired fixed-sign Hadamard transform (H), when enabled, applies to
  both column operands consumed by Wgrad.}
  \label{fig:qkv-ffn-fusion}
\end{figure}

Figure~\ref{fig:qkv-ffn-fusion} presents four execution panels:
QKV forward and backward, and FFN forward and backward. The serial schedule
writes BF16 intermediates to high-bandwidth
memory (HBM), reads them back to compute scales and layouts, and repeats that
cycle before the next GEMM. The fused schedule instead makes each producer
emit the exact scale metadata, row and column views, and saved BF16 state that
its consumers require.

For a fixed route, fusion preserves the defined numerical contract. It removes
intermediate traffic and exposes independent work to the scheduler. The scale stage is nevertheless
format-specific: \mxfp{} completes a local 32-value reduction; custom global
\nvfp{} starts a tensor-wide maximum early; and \localcta{} carries two tile
outer scales through the complete contraction and applies their product in the
GEMM epilogue.

The gated FFN illustrates why the boundary extends beyond a GEMM. It has gate
and up projections ($W_1$ and $W_3$), followed by a down projection ($W_2$).
Writing the $W_1$ and $W_3$ outputs to BF16, launching a separate
SiLU--multiply, then reading the result again to quantize the $W_2$ input
spends bandwidth between every large contraction. The fused path schedules the
paired projections together and emits the FP4 input to $W_2$ with the
SiLU--gate output. In backward, it combines the $W_2$ data gradient with the
SiLU derivative and produces the representations consumed by the $W_1$ and
$W_3$ data- and weight-gradient contractions. Exact BF16 values remain
available where autograd
requires them; fusion changes data movement, not the model definition.

QKV follows the same principle. The normalized input producer emits the
format-specific representations consumed by the grouped projection, and the
Q/K projection applies RoPE and the required layout. Backward combines inverse
layout, RoPE, QKV gradient production, and the adjacent RMSNorm reduction when
their data dependencies align. Residual--RMSNorm retains the exact residual
and row statistics while producing the next quantized representations.

Scale work is scheduled by format. \mxfp{} reductions finish with each local
tile. Custom global \nvfp{} begins the tensor maximum-absolute-value reduction
early so scale finalization can
overlap other work. \localcta{} removes that global dependency, but its two
outer scales must remain available through the full K reduction.
Figure~\ref{fig:format-execution-routes} summarizes these schedules.

\subsection{Depth-wise and operand-wise hybrids}

We call the \emph{depth-wise} route the 27/5 hybrid: it assigns complete
transformer blocks 1--27 to \localcta{} and blocks 28--32 to \mxfp{}.
Attention and FFN linears follow the
assigned block, so Fprop, Dgrad, and Wgrad never mix formats within one linear.
Both formats use orientation-consistent 2D weights and row-only gradient-data
SR; neither uses Hadamard preconditioning.

The \emph{operand-wise} hybrid instead chooses the format per contraction. A
controlled operator analysis using identical saved tensors found lower Dgrad
error for \localcta{} than for \mxfp{} in every measured layer and component.
This motivates replacing
\mxfp{} Dgrad with \localcta{} while retaining the \mxfp{} Fprop
and Wgrad path. The diagnostic and its scope are
reported in Section~\ref{sec:results} and
Appendix~\ref{app:dgrad-diagnostic}.

Concretely, every linear uses \mxfp{} for Fprop and Wgrad and
\localcta{} with row-data SR for Dgrad. One producer loads each BF16
gradient tile once and emits both the CTA-local row representation and the
MXFP4 column representation. The weight producer likewise loads BF16 $W$ once and
emits the MXFP4 forward view and the CTA-local Dgrad view.
The lower panel of Figure~\ref{fig:linear-operand-map} shows this
contraction-wise assignment for the plain-H16 variant.

We evaluate two Wgrad variants. The plain-H16 variant applies the same
16-value Hadamard mixing to $X$ and $dY$ with $D=I$, so it performs no sign
flips. The fixed-H32 variant applies the paired 32-value signed transform used
by the completed \mxfp{}+row-SR+fixed-H32 route. Neither variant transforms the
stored weight. Fprop, CTA-local Dgrad, weight encodings, row SR, output
projection, and loss retain the same numerical contract. Table~\ref{tab:gb200-throughput}
reports the two complete configurations; Appendix~\ref{app:implementation}
gives the sign construction.

Every main custom route retains a BF16 output projection and compiled regular cross
entropy. Appendix~\ref{app:output-head} summarizes the low-precision output
heads that did not improve the end-to-end throughput--numerical tradeoff.
\FloatBarrier

\section{Experimental Setup}

\subsection{Model, data, and training horizon}

The main workload is an 8B-parameter Llama-family decoder with RMSNorm,
grouped-query attention, rotary position embeddings (RoPE), and a gated
feed-forward network \cite{dubey2024llama3}. The sequence length is 8,192 and
the global batch is 512 sequences, or 4,194,304 tokens per update. A
160B-token run ends at step 38,147. Unless stated otherwise, the named
low-precision route applies to every body linear in all 32 transformer blocks.

All main custom routes use a BF16 output projection followed by
Inductor-compiled regular cross entropy. The optimizer is AdamW with a
2,000-step linear warmup to a peak learning rate of $3\times10^{-4}$; the rate
remains constant after warmup over the reported horizon. Training losses are
compared at matched optimizer steps. Each long-run recipe is represented by
one trajectory, so the comparisons do not estimate seed-to-seed variation.
Table~\ref{tab:base-training-recipe} summarizes the shared configuration; the
appendix records execution-specific details.

\begin{table}[htbp]
  \centering
  \footnotesize
  \caption{Shared Llama-8B training configuration.}
  \label{tab:base-training-recipe}
  \begin{tabularx}{\textwidth}{@{}lY@{}}
    \toprule
    Field & Effective contract \\
    \midrule
    Model & 32 blocks; width 4,096; FFN 14,336; 32 Q heads; 8 KV heads; head dimension 128; vocabulary 128,256 \\
    Position / norm & Llama-3.1 RoPE with base $500{,}000$, $8\times$ scaling, low/high frequency factors 1 and 4, and original length 8,192; RMSNorm $\epsilon=10^{-5}$ \\
    Data / tokenizer & Shuffled OLMo Mix 1124 \cite{olmo2025olmo2}; Meta-Llama-3.1-8B tokenizer \\
    Optimizer & Fused AdamW, $(\beta_1,\beta_2)=(0.9,0.95)$, $\epsilon=10^{-8}$, weight decay 0.1, global gradient-norm clip 1.0 \\
    LR schedule & 2,000-step linear warmup to $3\times10^{-4}$; constant thereafter over the reported horizon \\
    Batch / horizon & Global batch 512; 4,194,304 tokens/update; 38,147 updates; 160,000,114,688 scheduled tokens \\
    State / output & BF16 parameters and non-FP4 operations; FP32 master/optimizer state; BF16 output projection; execution-specific regular CE \\
    \bottomrule
  \end{tabularx}
\end{table}

\subsection{Recipe ledger}

Tables~\ref{tab:recipe-ledger} and~\ref{tab:recipe-state-ledger} state the
numerical recipe for every reported route. ``Row SR'' means stochastic
rounding only for the row-oriented gradient consumed by Dgrad. Activations,
weights, scales, and the separate Wgrad-oriented gradient representation
remain deterministic in every custom route labelled ``row SR.'' A Wgrad
Hadamard transform always acts on $X$ and $dY$ as a pair; it never transforms
the stored weight.

\begin{table}[htbp]
  \centering
  \footnotesize
  \caption{Per-contraction numerical recipe. 2D16 and 2D32 denote one
  orientation-consistent $16\times16$ or $32\times32$ weight encoding.
  NV denotes global NVFP4, LC denotes CTA-local NVFP4, and MX denotes MXFP4.
  The 27/5 hybrid changes format by depth; the operand hybrid changes it by
  contraction inside every block.}
  \label{tab:recipe-ledger}
  \begin{tabularx}{\textwidth}{@{}YYYYYY@{}}
    \toprule
    Route & Body assignment & Fprop & Dgrad & Wgrad & Weight representation(s) \\
    \midrule
    BF16 & blocks 1--32 & BF16 & BF16 & BF16 & BF16 master \\
    TE-native \nvfp{} in all blocks & blocks 1--32 & global NV & global NV & global NV + TE-managed H16 & TE 2D16 \\
    TE \nvfp{} with four final BF16 blocks & NV blocks 1--28; BF16 29--32 & NV 1--28; BF16 29--32 & NV 1--28; BF16 29--32 & NV+TE-managed H16 1--28; BF16 29--32 & TE 2D16 / BF16 \\
    Custom global \nvfp{} & blocks 1--32 & global NV & global NV & global NV & custom 2D16 \\
    \mxfp{} + row-SR & blocks 1--32 & MX & MX & MX & custom 2D32 \\
    \localcta{} & blocks 1--32 & localCTA & localCTA & localCTA & custom 2D16 \\
    \localcta{}+fixed H16 & blocks 1--32 & localCTA & localCTA & localCTA + paired fixed-sign H16 & custom 2D16 \\
    27/5 depth hybrid & LC blocks 1--27; MX 28--32 & LC 1--27; MX 28--32 & LC 1--27; MX 28--32 & LC 1--27; MX 28--32 & LC 2D16; MX 2D32 \\
    \mxfp{}+row-SR+fixed H32 & blocks 1--32 & MX & MX & MX + paired fixed-sign H32 & custom 2D32 \\
    Operand hybrid (plain H16) & every block & MX & localCTA & MX + paired plain H16 & MX 2D32 + LC 2D16 encodings \\
    Operand hybrid (fixed H32) & every block & MX & localCTA & MX + paired fixed-sign H32 & MX 2D32 + LC 2D16 encodings \\
    \bottomrule
  \end{tabularx}
\end{table}

\begin{table}[htbp]
  \centering
  \footnotesize
  \caption{Rounding and precision policy. Every row uses a BF16 output
  projection.}
  \label{tab:recipe-state-ledger}
  \begin{tabularx}{\textwidth}{@{}lYYY@{}}
    \toprule
    Route & Gradient rounding & Wgrad preconditioning & BF16 body blocks / output head \\
    \midrule
    BF16 & n/a & none & all / BF16 \\
    TE-native \nvfp{} in all blocks & TE gradient SR & TE-managed H16 on Wgrad & none / BF16 \\
    TE \nvfp{} + four final BF16 blocks & TE gradient SR & TE-managed H16 on NV Wgrad & last four / BF16 \\
    Custom global \nvfp{} & row data SR & none & none / BF16 \\
    \mxfp{} + row-SR & row data SR & none & none / BF16 \\
    \localcta{} & row data SR & none & none / BF16 \\
    \localcta{}+fixed H16 & row data SR & paired fixed-sign H16 on Wgrad $X,dY$ & none / BF16 \\
    27/5 depth hybrid & row data SR in both & none & none / BF16 \\
    \mxfp{}+row-SR+fixed H32 & row data SR & paired fixed-sign H32 on Wgrad $X,dY$ & none / BF16 \\
    Operand hybrid (plain H16) & LC row data SR & plain H16 ($D=I$) on Wgrad & none / BF16 \\
    Operand hybrid (fixed H32) & LC row data SR & paired fixed-sign H32 on Wgrad $X,dY$ & none / BF16 \\
    \bottomrule
  \end{tabularx}
\end{table}

``TE-managed'' means that Transformer Engine controls the H16 transform. Its
sign-state contract is distinct from the explicit fixed-sign contract used by
the custom kernels.

All custom training routes retain BF16 parameters and FP32 optimizer state. The final column
in Table~\ref{tab:recipe-ledger} describes the transient FP4 representation
consumed by Tensor Cores. The operand hybrid emits two encodings from one BF16
weight: MX 2D32 for Fprop and CTA-local 2D16 for Dgrad. Wgrad does not consume
the stored weight; it quantizes $X$ and $dY$ with MXFP4. This is a
deliberate contraction-specific recipe rather than two layouts of one shared
FP4 weight.

Every operand marked ``MX'' in Table~\ref{tab:recipe-ledger} uses
encode-centric E8M0 scales for activation and gradient blocks and the adaptive
2D32 weight rule in Section~\ref{sec:scale-contracts}. The scale exponent
itself is deterministic. ``Row SR'' applies to the E2M1 gradient values, not
to their E8M0 scale. This distinction is part of the recipe: scale direction
controls range and grid spacing, whereas data SR controls how a value is
selected between adjacent E2M1 levels.

Appendix~\ref{app:reproducibility} gives the checkpoint inclusion criteria and
additional reproducibility details.

\FloatBarrier
\subsection{End-to-end performance measurements}

The performance probes use the same accelerator type, sequence length, global
batch, seed, BF16 output projection, compiled CE, warmup rule, metric window,
and per-GPU timing records. Each route uses its selected local batch,
activation-checkpoint policy, and distributed schedule, so the panel compares
complete end-to-end configurations rather than one isolated implementation
change. Tables report cross-device aggregates. Local batch (LB) and gradient
accumulation (GA) remain explicit because they affect occupancy, memory
lifetime, and the exposed output-layer cost.

We use the dense-model training FLOP convention implemented by TorchTitan. For
total parameter count $N$, embedding parameter count
$N_{\mathrm{emb}}$, number of layers $L$, query-head count $H$, head dimension
$d_h$, and sequence length $S$, it gives
\begin{equation}
  F_{\mathrm{model/token}}
  =6\bigl(N-N_{\mathrm{emb}}\bigr)+12LHd_hS.
  \label{eq:model-flops}
\end{equation}
The first term counts the forward and backward dense parameter operations;
the second counts the attention contractions without charging recomputation.
Here $N=8{,}030{,}261{,}248$, $N_{\mathrm{emb}}=525{,}336{,}576$, $L=32$,
$H=32$, $d_h=128$, and $S=8{,}192$, giving
$F_{\mathrm{model/token}}=57{,}914{,}449{,}920$ FLOPs/token.

For a route with per-device token rate $R$, we report BF16 model FLOP
utilization (BF16 MFU):
\begin{equation}
  \mathrm{MFU}_{\mathrm{BF16}}=100\frac{F_{\mathrm{model/token}}R}
  {F_{\mathrm{peak,BF16}}},
  \label{eq:mfu}
\end{equation}
using Equation~\ref{eq:model-flops} and the hardware's maximum dense BF16
Tensor Core throughput. NVIDIA specifies 360 PFLOP/s of sparse FP16/BF16
throughput for a 72-GPU GB200 NVL72 and states that dense throughput is half
the sparse value. Dividing by 72 GPUs and by two gives the maximum dense BF16
throughput used here, $F_{\mathrm{peak,BF16}}=2.5\times10^{15}$ FLOP/s/GPU,
or 2,500 BF16 TFLOP/s/GPU \cite{nvidia2025blackwell}. Tokens/s/GPU is the
directly observed metric. We recompute BF16 MFU from each unrounded throughput
aggregate with this peak. For an FP4 route, BF16 MFU is BF16-equivalent model
work divided by the BF16 peak; it measures effective training throughput, not
literal FP4 Tensor Core occupancy. It can therefore exceed 100\%.
Recomputing from the rounded token rate printed in a table can differ in the
last decimal place.

\subsection{Training and checkpoint evaluation}

Training curves use exact logged optimizer steps. Smoothed curves use an
exponential moving average only for visualization. Custom-route endpoints are
raw observations; Appendix~\ref{app:reproducibility} defines the BF16 and
TE-native reference curves.

Validation uses a fixed external stream constructed from an 82/18 DataComp-LM
(DCLM) and OLMo-without-DCLM source mixture
\cite{li2024datacomplm,olmo2025olmo2}. It contains 768
sequences and 6,291,456 next-token targets. Documents are deduplicated, mixed,
and shuffled before packing so one GPU worker cannot receive examples from
only one source or contiguous region of source data. Every checkpoint consumes
the same tokenized sequences. We
call this an external validation stream rather than a held-out split because
zero overlap with every historical training trajectory cannot be proven.
Before route comparison, the converted-checkpoint evaluator and native
TorchTitan model must agree at the full 8,192-token context, and a late BF16
checkpoint must improve over an early one by more than five sequence-level
standard errors.

Every downstream comparison uses the literal step-38,000 checkpoint and BF16
inference. Converted models must match TorchTitan's RoPE and RMSNorm semantics
before scoring. The evaluation panel follows the four-task subset used by the
reference NVFP4 protocol: MMLU, HellaSwag, WinoGrande, and ARC-Challenge
\cite{hendrycks2021mmlu,gu2024olmes,nvidia2025nvfp4}. Shot counts and reported metrics appear with the
results.

\section{Results}
\label{sec:results}

\subsection{What throughput do the end-to-end routes achieve?}

The fastest custom path, \mxfp{} with row-gradient SR, reaches 37,876.8
tokens/s/GPU, compared with
18,812.0 for BF16 and 27,622.0 for TE-native \nvfp{} in all blocks.
Table~\ref{tab:gb200-throughput} reports the complete same-accelerator panel.
Every row uses sequence length 8,192, global batch 512, a BF16 output
projection, compiled regular CE, and a cross-device median over steps 40--100.
BF16 and TE use local batch size 2 with eight gradient-accumulation steps
(LB2/GA8); the custom paths use local batch size 4 with four
gradient-accumulation steps (LB4/GA4).

\begin{table}[H]
  \centering
  \caption{GB200 throughput with compiled regular CE. Values are cross-device
  medians over steps 40--100 for each route's selected batch geometry. BF16
  uses the selective-8 setting, which checkpoints every eighth transformer
  block; all other rows use selective-32, which checkpoints every 32nd block.
  BF16 MFU uses NVIDIA's 2,500-TFLOP/s/GPU maximum dense BF16 throughput:
  180 PFLOP/s dense for 72 GPUs \cite{nvidia2025blackwell}.}
  \label{tab:gb200-throughput}
  \begin{adjustbox}{max width=\textwidth}
  \begin{tabular}{ll@{\qquad}rrr}
    \toprule
    Route & LB/GA & Tokens/s/GPU & BF16 MFU & Median HBM \\
    \midrule
    BF16 & 2/8 & 18,812.00 & 43.58\% & 101.19 GiB \\
    TE-native \nvfp{} in all blocks & 2/8 & 27,622.00 & 63.99\% & 125.80 GiB \\
    TE \nvfp{} + four final BF16 blocks & 2/8 & 27,077.50 & 62.73\% & 126.03 GiB \\
    Custom global \nvfp{} & 4/4 & 32,490.50 & 75.27\% & 155.63 GiB \\
    \mxfp{} + row-SR & 4/4 & 37,876.75 & 87.74\% & 154.43 GiB \\
    \localcta{} & 4/4 & 33,243.76 & 77.01\% & 115.52 GiB \\
    \localcta{} + fixed H16 & 4/4 & 31,882.61 & 73.86\% & 114.89 GiB \\
    27/5 depth hybrid & 4/4 & 33,120.50 & 76.73\% & 126.21 GiB \\
    Operand hybrid + plain H16 & 4/4 & 34,782.50 & 80.58\% & 154.92 GiB \\
    Operand hybrid + fixed H32 & 4/4 & 35,085.50 & 81.28\% & 150.89 GiB \\
    \mxfp{} + row-SR + fixed H32 & 4/4 & 37,247.00 & 86.29\% & 151.02 GiB \\
    \bottomrule
  \end{tabular}
  \end{adjustbox}
\end{table}

Under the same 32-GPU geometry with local batch size 4 and four gradient
accumulation steps (LB4/GA4), the
\mxfp{}+row-SR+fixed-H32 route reaches 98.34\% of the row-SR-only \mxfp{} throughput.
Additional measurement details appear in
Appendix~\ref{app:implementation}.

\subsection{How closely do the recipes track BF16 training loss?}

\begin{figure}[tbp]
  \centering
  \includegraphics[width=0.98\textwidth]{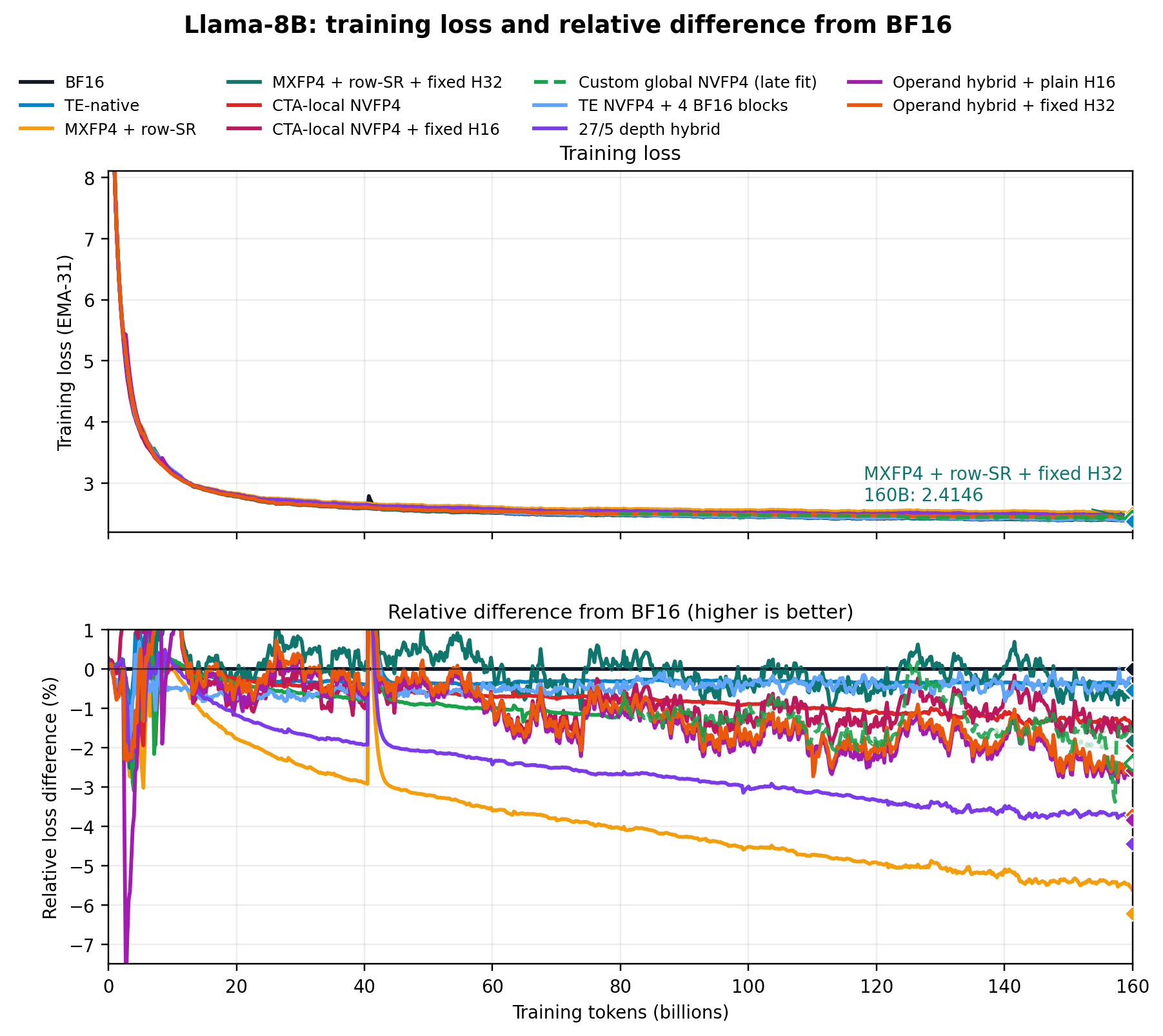}
  \caption{Llama-8B training loss and relative difference from the BF16
  reference. Continuous histories use EMA-31, and diamonds mark the final
  plotted endpoint. Custom-global-\nvfp{} has regular observations through
  step 19,000, but its dense later history was not retained. The upper panel
  therefore shows the 74 exact later observations recovered from training
  logs, including the terminal diamond; the lower panel shows the 18
  observations recorded at the same step as BF16. The dashed
  custom-global-\nvfp{} line is a smoothed visual guide through those sparse
  observations, not additional measurements. Appendix~\ref{app:reproducibility}
  records the surviving data sources.}
  \label{fig:llama8b-loss}
\end{figure}

Figure~\ref{fig:llama8b-loss} compares the training trajectories. The top panel
shows an exponential moving average over 31 logged observations (EMA-31). The
bottom panel follows the
relative-difference convention used by NVIDIA
\cite{nvidia2025nvfp4}:
\begin{equation}
  \Delta_{\mathrm{BF16}}(t)=100
  \frac{L_{\mathrm{BF16}}(t)-L_{\mathrm{route}}(t)}
       {L_{\mathrm{BF16}}(t)}.
  \label{eq:relative-loss}
\end{equation}
Zero is the BF16 reference and a negative value means the low-precision route
has higher loss. Reported comparisons use exact shared logged steps; no
reported loss value is interpolated. The dashed custom-global-NVFP4 line is the
display-only fit described below.

\begin{figure}[tbp]
  \centering
  \includegraphics[width=0.98\textwidth]{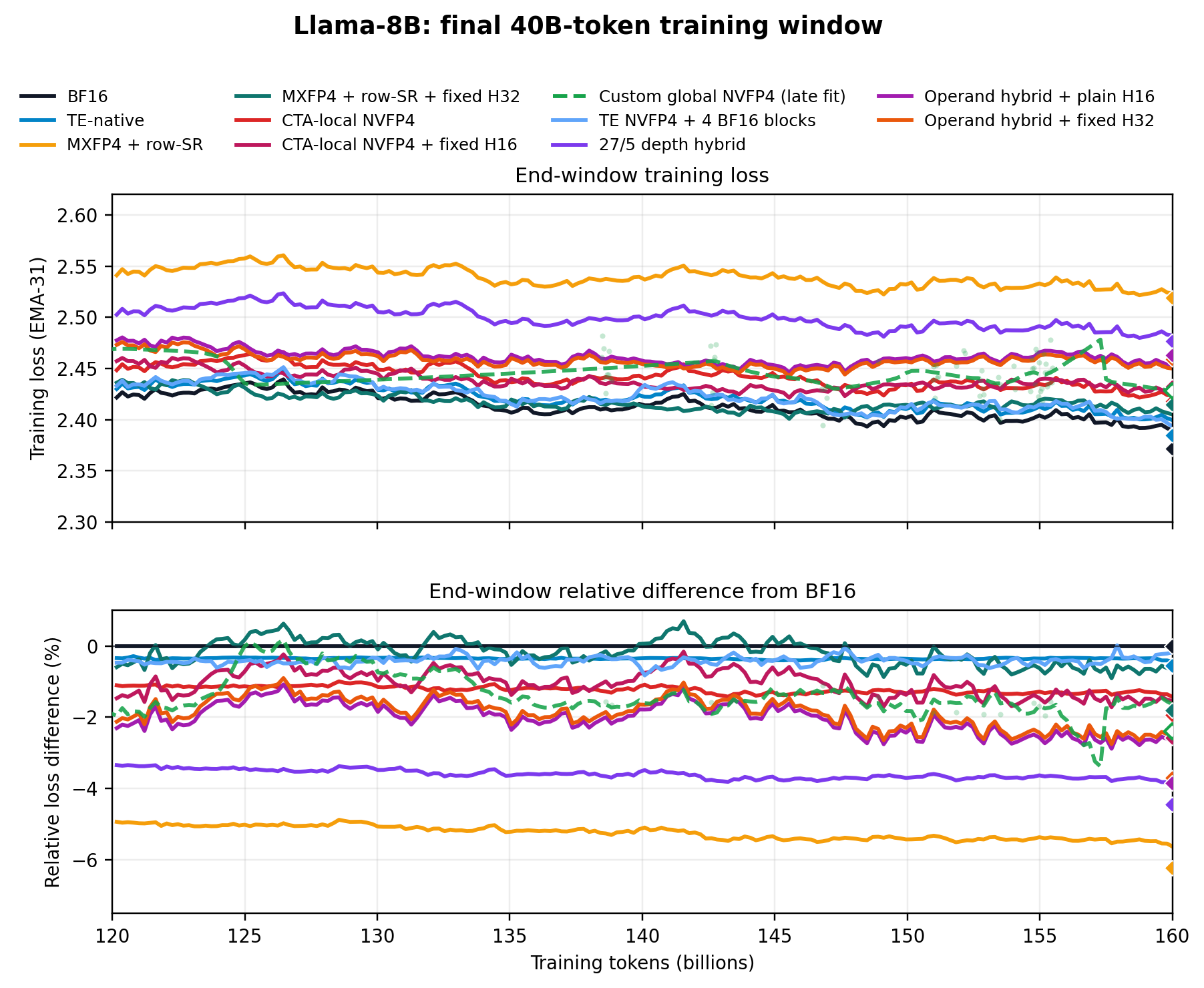}
  \caption{Training loss from 120B to 160B tokens. Diamonds mark exact logged
  endpoints. As in Figure~\ref{fig:llama8b-loss}, the late
  custom-global-\nvfp{} markers are exact observations and the dashed curve is
  a smoothed visual guide, not additional measurements.}
  \label{fig:llama8b-loss-zoom}
\end{figure}

Figure~\ref{fig:llama8b-loss-zoom} expands the final 40B-token window. It shows
the late loss separation at the same scale without extrapolating any route.

The TE route with four final BF16 blocks and the fixed-H32 operand hybrid both
complete the scheduled training horizon.

\begin{table}[H]
  \centering
  \caption{Raw last logged minibatch loss at step 38,140 (159.97B tokens).
  ``Loss excess'' is relative to the unadjusted BF16 observation at that step.}
  \label{tab:training-endpoints}
  \begin{tabularx}{\textwidth}{Yrr}
    \toprule
    Route & Raw endpoint loss & Loss excess vs BF16 \\
    \midrule
    BF16 & 2.364637 & 0.00\% \\
    TE \nvfp{} + four final BF16 blocks & 2.385181 & 0.87\% \\
    TE-native \nvfp{} & 2.396353 & 1.34\% \\
    \mxfp{} + row-SR + fixed H32 & 2.414578 & 2.11\% \\
    \localcta{} & 2.417708 & 2.24\% \\
    Custom global \nvfp{} & 2.428200 & 2.69\% \\
    \localcta{} + fixed H16 & 2.432642 & 2.88\% \\
    Operand hybrid + fixed H32 & 2.459305 & 4.00\% \\
    Operand hybrid + plain H16 & 2.462673 & 4.15\% \\
    27/5 depth hybrid & 2.476842 & 4.75\% \\
    \mxfp{} + row-SR & 2.519150 & 6.53\% \\
    \bottomrule
  \end{tabularx}
\end{table}

At step 38,140, the TE route with four final BF16 blocks is the closest
quantized trajectory to the raw BF16 observation, with 0.87\% excess, followed
by TE-native at 1.34\%. \mxfp{} with row-SR and fixed H32 is the closest custom
route at 2.11\%; it is 0.76\% above TE-native. The fixed-H32 operand hybrid
ends 1.85\% above the pure \mxfp{}-H32 route and 4.00\% above BF16.

To compare error in the backward input-gradient calculation (Dgrad), we used
tensors captured at step 28,000 from the \mxfp{}+row-SR+fixed-H32 run. We
recomputed Dgrad with \localcta{} and with \mxfp{}, using BF16 as the
reference. \localcta{} had lower Dgrad error energy in all 288
layer--component comparisons; the geometric-mean
$E_{\mathrm{MX}}/E_{\mathrm{localCTA}}$ ratio was 1.292. The completed
plain-H16 operand hybrid nevertheless ends 1.99\% above \mxfp{}+row-SR+fixed H32 in
training loss. Appendix~\ref{app:dgrad-diagnostic} gives the capture and
aggregation contract.

\subsection{Validation loss}

\begin{figure}[tbp]
  \centering
  \includegraphics[width=0.98\textwidth]{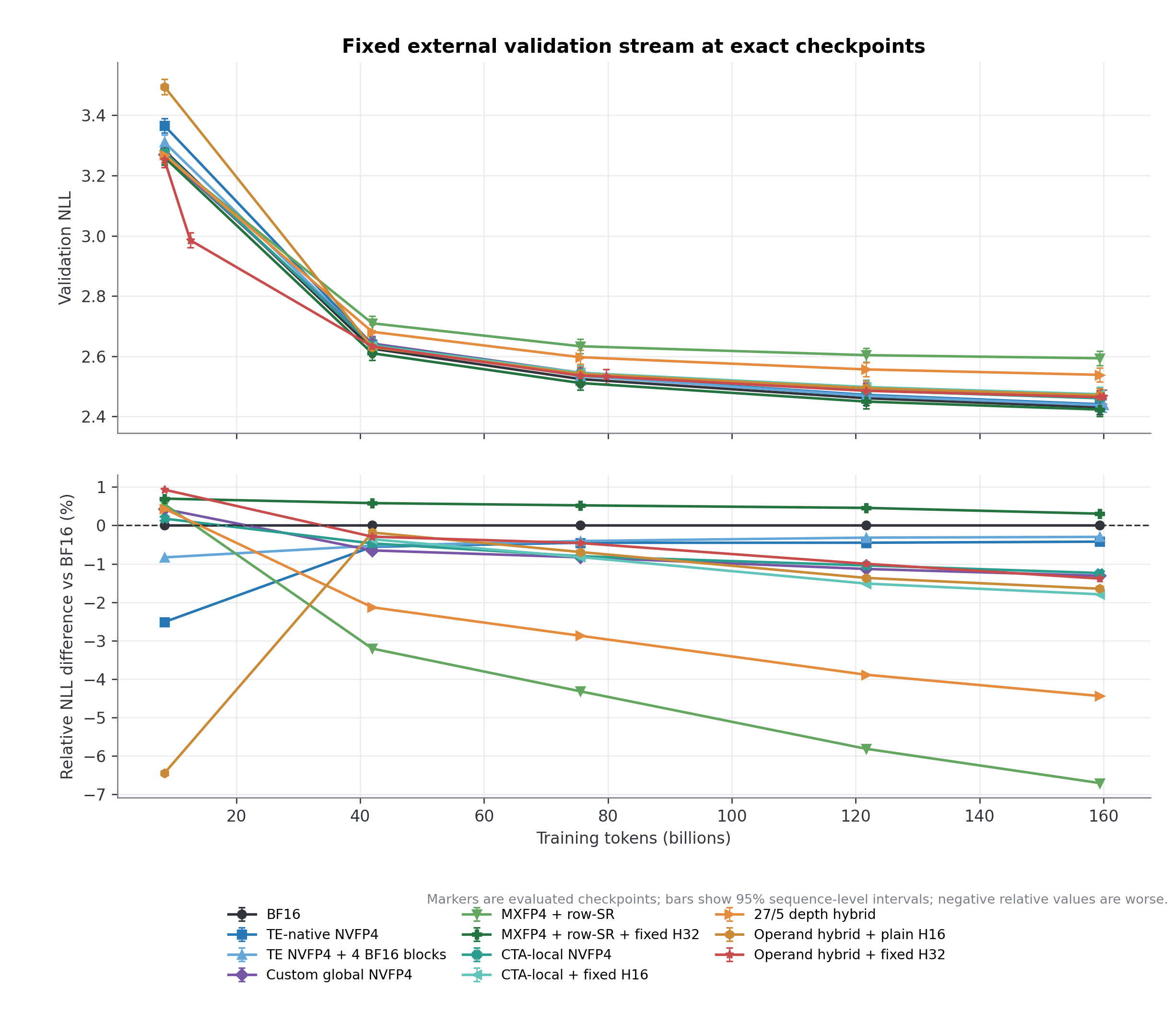}
  \caption{Validation NLL on the fixed external stream at every available exact
  checkpoint. Error bars are 95\% sequence-level intervals. The lower panel
  reports $100(L_{\mathrm{BF16}}-L_{\mathrm{route}})/L_{\mathrm{BF16}}$ at
  exact shared steps, so negative values are worse than BF16. Lines connect
  evaluated checkpoints only; no value is interpolated or extrapolated.}
  \label{fig:llama8b-validation}
\end{figure}

Figure~\ref{fig:llama8b-validation} evaluates every available exact checkpoint
on one fixed external stream. We report mean negative log-likelihood (NLL),
for which lower values are better. At the common step-38,000 cut, \mxfp{} with fixed H32 has
the lowest observed validation NLL, 2.422804. BF16 records 2.430206, the TE
route with four final BF16 blocks 2.437475, and TE-native 2.440488. The
fixed-H32 operand hybrid records 2.463842, compared with 2.460331 for
\localcta{} and 2.462074 for custom global \nvfp{}. Relative to BF16, the
relative NLL differences $100(L_{\mathrm{BF16}}-L_{\mathrm{route}})/L_{\mathrm{BF16}}$
for \mxfp{}-H32, four-final-BF16-block, and operand-H32 are
$+0.305\%$, $-0.299\%$, and $-1.384\%$, respectively; positive values
indicate lower NLL than BF16. These comparisons cover complete recipes from
one training trajectory each; they do not isolate the effect of H32 or average
over repeated seeds.

The 27/5 depth hybrid is evaluated at the five common cuts: steps 2,000,
10,000, 18,000, 29,000, and 38,000. At step 38,000 its validation NLL is
2.538116, or 4.440\% higher than matched BF16.

\FloatBarrier

\subsection{How do the routes compare on downstream tasks?}

The common panel evaluates all eleven routes with a complete exact step-38,000
checkpoint. Every route uses BF16 inference and the same MMLU, HellaSwag,
WinoGrande, and ARC-Challenge protocol.

\begin{table}[htbp]
  \centering
  \footnotesize
  \caption{Exact step-38,000 downstream scores (\%). MMLU and WinoGrande
  report accuracy; HellaSwag and ARC-Challenge report normalized accuracy.
  Shot counts are 5, 10, 5, and 25, respectively. Every route uses BF16
  inference after exact converted/native parity. Bold marks the highest
  observed score in each column.}
  \label{tab:downstream-step38}
  \begin{adjustbox}{max width=\textwidth}
  \begin{tabular}{lrrrr}
    \toprule
    Route & MMLU & HellaSwag & WinoGrande & ARC-Challenge \\
    \midrule
BF16 & \textbf{28.64} & 67.50 & 62.12 & 42.06 \\
TE-native \nvfp{} & 25.19 & 66.62 & 63.14 & 42.49 \\
TE \nvfp{}, four final BF16 blocks & 26.45 & 66.97 & 61.01 & 43.26 \\
Custom global \nvfp{} & 27.38 & 65.81 & 63.85 & 41.72 \\
\mxfp{} + row-SR & 25.94 & 60.39 & 60.69 & 37.54 \\
\mxfp{} + row-SR + fixed H32 & 25.47 & \textbf{68.82} & 63.61 & \textbf{44.54} \\
\localcta{} & 25.54 & 66.32 & 62.75 & 42.66 \\
\localcta{} + fixed H16 & 23.74 & 65.08 & 63.06 & 43.34 \\
27/5 depth hybrid & 25.16 & 62.94 & 61.17 & 39.08 \\
Operand hybrid + plain H16 & 24.40 & 66.47 & 62.98 & 42.92 \\
Operand hybrid + fixed H32 & 23.81 & 66.80 & \textbf{64.40} & 42.15 \\
\bottomrule

  \end{tabular}
  \end{adjustbox}
\end{table}

The task ordering is not uniform. BF16 has the highest MMLU score, the
fixed-H32 operand hybrid has the highest WinoGrande score, and
\mxfp{}+row-SR+fixed H32 has the highest HellaSwag and ARC-Challenge scores.
Relative to that pure \mxfp{}-H32 route, the fixed-H32 operand hybrid differs
by $-1.66$, $-2.02$, $+0.79$, and $-2.39$ points on MMLU, HellaSwag,
WinoGrande, and ARC-Challenge, respectively. The four-final-BF16-block TE
route differs from BF16 by $-2.19$, $-0.53$, $-1.10$, and $+1.19$ points.
These are observations from one checkpoint per trajectory, not repeated-seed
estimates.

\begin{figure}[htbp]
  \centering
  \includegraphics[width=0.98\textwidth]{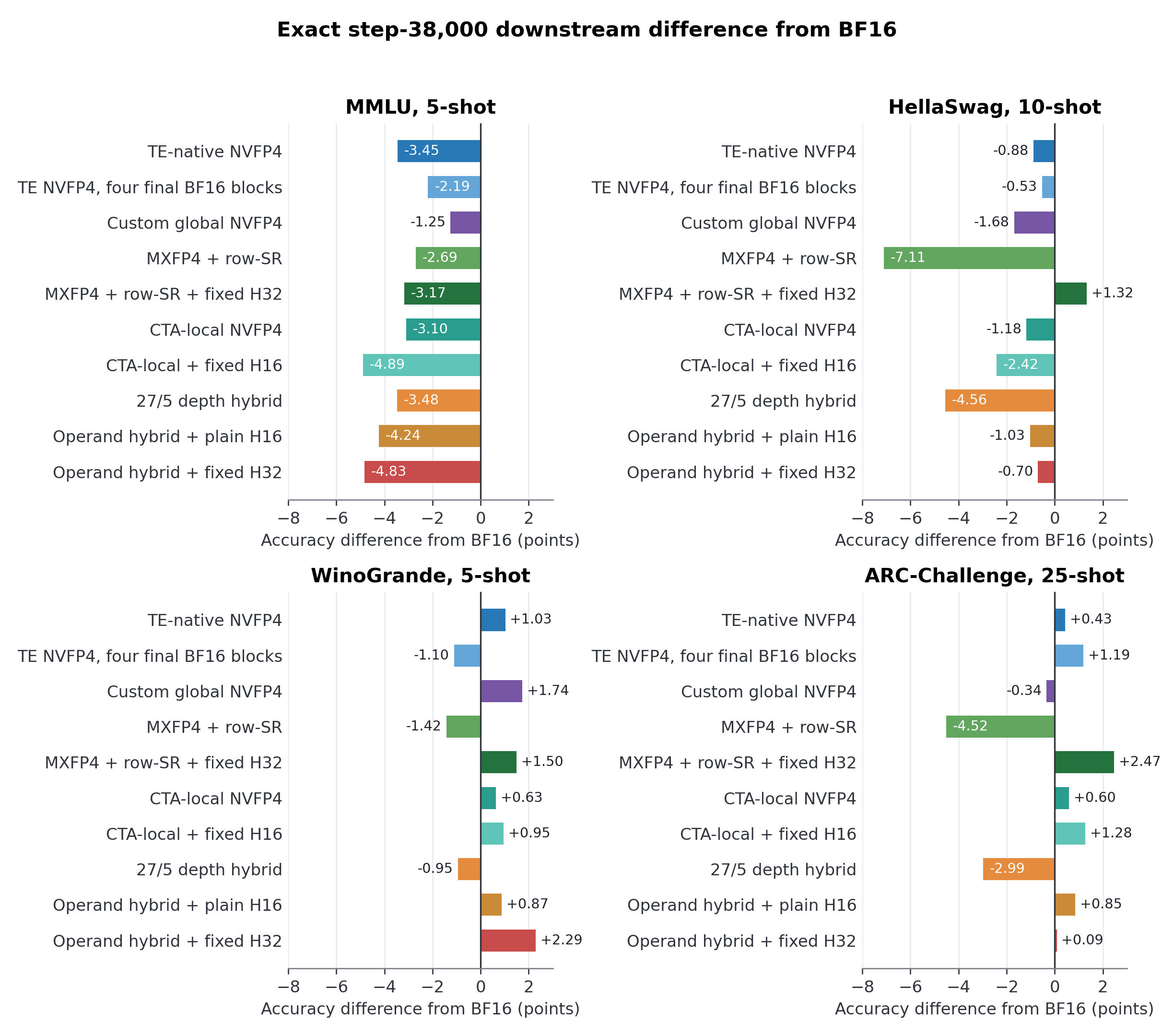}
  \caption{Downstream reported-score difference from BF16 at exact step 38,000.
  Positive values exceed the BF16 score on that task. All panels use the same
  percentage-point scale. The absolute scores are reported in
  Table~\ref{tab:downstream-step38}.}
  \label{fig:downstream-delta}
\end{figure}
\FloatBarrier

\section{Discussion}

The measured TE routes track BF16 closely: TE-native and the
four-final-BF16-block route end 1.34\% and 0.87\% above the raw BF16 loss.
Their common probes reach 27.6K and 27.1K tokens/s/GPU. A separate
same-accelerator probe of the custom global \nvfp{} route reaches 32.5K
tokens/s/GPU; its long trajectory ends 2.69\% above BF16. TE-native \nvfp{}
and custom global \nvfp{} both use the tensor-global NVFP4 scale hierarchy;
the four-final-BF16-block route uses that hierarchy in its first 28 blocks.
The panel does not isolate the cost
of scaling from implementation and batch geometry, but it exposes the systems
constraint: a tensor-global maximum-absolute-value reduction creates a
completion dependency that the execution path must hide or pay.

Hadamard preconditioning is not uniformly beneficial across scale strategies.
The fixed-H32 \mxfp{} recipe retains 98.34\% of row-SR-only \mxfp{}
throughput while moving the raw endpoint excess from 6.53\% to 2.11\% and
validation NLL from 2.593146 to 2.422804. The fixed-H16 CTA-local recipe is
slower than CTA-local without the transform and has higher endpoint and
validation loss. These are comparisons between complete trajectories, not
experiments that isolate the effect of Hadamard preconditioning alone.

Neither tested hybrid dominates the strongest pure custom route. The
depth-wise hybrid and both operand-wise hybrids are slower and have higher
endpoint and validation loss than \mxfp{} with row-SR and fixed H32. A
controlled replay on identical saved tensors still finds lower Dgrad error for
CTA-local in all 288 tested layer--projection comparisons. The completed hybrid
trajectories do not preserve that operator-level advantage in endpoint or
validation loss.

\section{Conclusion and Further Work}

Format-aware fusion makes the complete FP4 producer--consumer boundary the
unit of optimization. The TE routes give the closest observed training-loss
tracking, with measured tensor-global execution paths at 27.1--27.6K
tokens/s/GPU. Among the custom routes, \mxfp{} with row-gradient SR and
fixed-sign H32 Wgrad preconditioning reaches 37.2K tokens/s/GPU, 86.3\%
BF16 MFU, a 2.11\% raw
endpoint-loss excess, and the lowest observed validation NLL on the fixed
external data stream. Its
downstream ranking remains task dependent. BF16 records the highest MMLU
score, the fixed-H32 operand hybrid the highest WinoGrande score, and
\mxfp{} with row-SR and fixed H32 the highest HellaSwag and ARC-Challenge
scores. In this panel, late training loss and downstream accuracy therefore do
not have a monotonic ordering; training loss alone does not rank the downstream
quality of these complete low-precision recipes.

Neither the tested depth-wise nor operand-wise hybrid improves the combined
speed--quality point of the strongest pure custom route. The opposite outcomes
of Hadamard preconditioning under \mxfp{} and CTA-local scaling motivate
choosing the group of values sharing each scale, the rounding rule, and the
preconditioner jointly. A complementary direction is to change the scale
representation itself. UE5M3 FP4 pairs E2M1 payloads with an unsigned E5M3
block scale and has supported a recipe with periodic tensor-scale updates,
selective SR, and no Hadamard preconditioning on a separate model and training
setup \cite{hu2026ue5m3}. Combining wider unsigned block scales with a
CTA-local producer schedule, or with delayed tensor scaling and global outer
scales, is a natural next step.

\section*{Generative-AI Disclosure}

Generative AI tools assisted with code development, experiment orchestration,
analysis, figure generation, and drafting and editing this report. The author
reviewed and verified the code, experiments, numerical results, figures,
citations, and final text and takes responsibility for the content.

\bibliographystyle{plainnat}
\bibliography{references}

\appendix
\section{Implementation and Validation Notes}
\label{app:implementation}

The kernels specialize the production tensor shapes and define buffer
lifetimes across eager execution, recomputation, and CUDA Graph replay. QKV and
FFN producers emit both quantized orientations. Residual--RMSNorm preserves the
BF16 residual and exact row statistics required by backward. Each path is
checked against a higher-precision reference for outputs, input gradients,
weight gradients, relative RMS error, and zero fractions before model timing.

Row- and column-oriented scales are distinct numerical state, not
interchangeable layout metadata. Tensors returned to autograd cannot alias a
workspace whose lifetime ends with the producing stream. Each delayed FP4
prefetch buffer has one owner and is released exactly once, preventing
asynchronous work from outliving its storage. Distributed measurements use
per-GPU timing and dispatch evidence because the slowest GPU sets the step
time.

\subsection{Weight orientation and the fixed sign pattern}

The custom global NVFP4 path creates one 2D16 numerical weight
encoding and exposes it in both consumer layouts. The native 2D32 \mxfp{} path
follows the same orientation-consistency rule.

The signed-Hadamard paths use the fixed literal \texttt{0x00002817}. Under the
custom kernel's convention, a set bit denotes $+1$ and a clear bit denotes
$-1$. Bits 0--15 encode the same 16-entry sign vector that the pinned TE
implementation stores explicitly. The H32 implementation applies this same
16-sign pattern separately to two 16-value halves, then combines the halves
with the final 32-value Hadamard add/subtract stage. Thus H32 names the
32-value transform width, not a 32-bit random mask: the reported H32 transform
repeats one signed-H16 diagonal and does not interpret the literal's upper
zero bits as 16 additional signs. No signs are drawn during training. The
construction remains orthogonal and preserves the unquantized Wgrad dot
product because the identical transform is paired on $X$ and $dY$. The
experiments do not compare this fixed vector with newly sampled signs and do
not establish that the particular diagonal is optimal.

Compilation and first-step kernel generation are excluded from steady-state
probe windows. Each measurement records device count, local batch, gradient
accumulation, seed, loss implementation, source revision, and metric window.
Each distributed worker checkpoints its stochastic-rounding seed and counter
offset so that a resumed run does not reuse random-number draws.

Table~\ref{tab:b300-snapshots} gives secondary production measurements. All
three use a BF16 output projection and compiled regular CE. Their batch
geometries and measurement windows differ, so the table does not isolate one
implementation change.

\begin{table}[H]
  \centering
  \caption{Compiled-CE B300 production snapshots.
  BF16 MFU uses the B300's 2,250 TFLOP/s/GPU dense BF16 peak
  \cite{nvidia2026hgx}.}
  \label{tab:b300-snapshots}
  \begin{tabularx}{\textwidth}{Ylrrl}
    \toprule
    Route & LB/GA & Tokens/s/GPU & BF16 MFU & HBM/GPU \\
    \midrule
    \mxfp{} + row-SR & 4/4 & 39,306.59 & 101.17\% & 156.91--157.15 GiB \\
    \localcta{} & 8/2 & 36,728.43 & 94.54\% & 211.05--211.16 GiB \\
    27/5 depth hybrid & 8/2 & 37,394.00 & 96.25\% & 228.28--233.42 GiB \\
    \bottomrule
  \end{tabularx}
\end{table}

\section{Low-Precision Output Head: Negative Result}
\label{app:output-head}

This appendix records an engineering screen for which the exact low-precision
forward format was not preserved. The vocabulary projection has a large
forward GEMM and is therefore an attractive FP4 target. In the complete
trainer, however, the head also owns cross entropy, the hidden-state gradient,
the weight gradient, quantization, and saved-tensor lifetime. Accelerating one
GEMM did not accelerate that whole boundary.

MXFP8 below denotes an eight-bit microscaling format used for one
weight-gradient configuration.

\begin{table}[H]
  \centering
  \caption{Historical whole-trainer output-head screen. Deltas are paired
  within one experiment. The complete profiler metadata was not preserved, so
  rates are rounded and are not compared with the primary performance panel.}
  \label{tab:head-negative}
  \begin{tabularx}{\textwidth}{YrrY}
    \toprule
    Forward / hidden-state gradient / weight gradient & Tokens/s/GPU & Difference & Outcome \\
    \midrule
    BF16 / BF16 / BF16 & $\sim$34.00K & baseline & BF16 control \\
    lowp$^{\dagger}$ / BF16 / BF16 & $\sim$33.95K & $\sim-0.16\%$ & Loss differed from control \\
    lowp$^{\dagger}$ / BF16 / MXFP8 & $\sim$33.91K & $\sim-0.27\%$ & Loss and gradients differed \\
    \bottomrule
  \end{tabularx}
\end{table}

$^{\dagger}$ The experiment record preserves the backward choices and paired
rates but not the exact low-precision forward format. We therefore retain the
literal ``lowp'' label rather than assigning a more specific format after the
fact.

Both selective low-precision variants are slower than their matched BF16
control. The preserved record also says that neither reproduced the BF16
control loss, and that the MXFP8 weight-gradient configuration additionally
changed gradients; it
does not retain a publication-ready error magnitude, so we do not rank the
size of those numerical differences. These results support the selected
design: a BF16 output projection with compiled CE.

\section{Reproducibility}
\label{app:reproducibility}

\subsection{Report inputs}

The main training and performance tables and figures are generated from
checked-in data snapshots. The publication data directory contains the
training-loss series, same-accelerator performance aggregates, checkpoint
inventories, evaluation results, plotting programs, and a SHA-256 ledger.
Figure generators verify their declared inputs before writing an output. The
machine-readable recipe ledger records the route, precision policy, rounding
and preconditioning choices, batch geometry, seed when recovered, checkpoint
horizon, and evidence status for every experiment discussed in the paper.
Shared model, data, and optimizer fields are stored separately so that an
incomplete historical row cannot silently inherit a value from a newer run.

For display consistency, the historical BF16 and TE-native training-loss
curves in Figure~\ref{fig:llama8b-loss} use fixed offsets inferred from 41
matched steps in seed-42 controls. The raw series and adjustment remain in the
publication data.

The long-run curves combine only explicitly identified execution segments.
The \localcta{}+fixed-H16 and operand-hybrid trajectories each cross a
documented full-state resume boundary. Their manifests identify the source and
continuation ranges; no overlapping step is counted twice. Training-loss
observations for custom global NVFP4 are regular through step 19,000. The
configured Weights \& Biases run is no longer returned by the service, and the
dense local CSV named by the historical manifest was not present in any
recovered or released tree. Seventy-four exact observations survive after
step 19,000 in fragmented training logs. The largest missing sequence of
expected ten-step records runs from step 24,710 through step 33,000. The
dashed late line is a visual LOWESS fit through the surviving observations,
evaluated at the BF16 plotting steps. It is never used to compute a reported numeric
endpoint, table entry, or statistic.

\subsection{Checkpoint and evaluator contract}

Training comparisons use exact shared steps. Custom-route values are raw
observations. Downstream comparisons use literal step-38,000
checkpoints; a terminal, nearby, or checkpoint from a different training run
is never substituted. Distributed checkpoints must contain all expected
shards and pass conversion checks before evaluation.

The evaluator must reproduce the training model's RoPE and RMSNorm semantics.
The Llama-3.1 8B training configuration uses base $500{,}000$ and the
long-context RoPE scaling parameters $(8,1,4,8192)$. An earlier evaluator
disabled this scaling in both the converted-checkpoint evaluator and native
TorchTitan reference. Agreement between those two implementations therefore
did not establish agreement with the training model. Results from that
evaluator are excluded.

An earlier validation stream used the correct model semantics but failed a
data-source robustness check: apparent route ordering changed when examples
came from different contiguous regions of the source datasets. Those values
are also excluded. The replacement stream was constructed afresh from
deterministic windows in the 90--95\% physical sample range of the DCLM and
OLMo-without-DCLM components. Documents are deduplicated by raw-text SHA-256,
combined at an 82/18 token ratio, globally hash-shuffled, and only then packed
and sharded. Thus no GPU worker is restricted to one source or contiguous
region of source data.

The replacement fixed validation stream contains 768 sequences of 8,193 tokens, yielding
6,291,456 scored next-token targets. It is tokenized once with the pinned
Llama-3.1 tokenizer and reused byte-for-byte for every checkpoint. We call it
a \emph{fixed external validation stream}, not a held-out split, because zero
document overlap across every historical training trajectory has not been
proven. Its source ledger contains 4,597 documents with
4,597 unique raw-text hashes. Before reporting route comparisons, the
corrected checkpoint evaluator must agree exactly with the native TorchTitan
model at the full 8,192-token context, and a late BF16 checkpoint must improve
over an early BF16 checkpoint by more than five sequence-level standard
errors. Both checks pass. The converted checkpoint and native TorchTitan model
agree exactly at full context for the step-2,000 and step-29,000 BF16
checkpoints. Their validation NLL values are 3.283272 and 2.460520,
respectively; the paired late-minus-early difference is $-0.822753$ with
sequence-level standard error 0.003708.

Downstream evaluation uses BF16 inference and one evaluator configuration for
all routes. MMLU is 5-shot, HellaSwag 10-shot, WinoGrande 5-shot, and
ARC-Challenge 25-shot; HellaSwag and ARC report normalized accuracy. A route
enters the common panel only after the exact checkpoint and corrected model
semantics are verified.

\subsection{Historical execution identity}

Historical experiments are bound to the code, submodule, kernel asset,
checkpoint, and output identities recovered for that execution. The current
source and runtime tips are not assigned retroactively to an older run. Where
a field was not recovered, the ledger records it as unknown rather than
guessing the nearest commit. Complete identifiers and content digests are in
the publication data ledger rather than the narrative.

The custom global NVFP4 path uses one orientation-consistent 2D16 weight
encoding. The code used by the completed \mxfp{}+row-SR+fixed-H32 run is
content-verified in its run manifest. The \localcta{}+fixed-H16 and
operand-wise plain-H16 trajectories each join verified execution segments at
an exact full-state checkpoint. Their manifests record implementation changes,
segment boundaries, and checkpoint state.

\subsection{Controlled Dgrad diagnostic}
\label{app:dgrad-diagnostic}

The operand-format diagnostic uses tensors from the step-28,000 checkpoint of
the completed \mxfp{}+row-SR+fixed-H32 trajectory. It computes Dgrad with
\mxfp{} and \localcta{} against the corresponding BF16 result. The diagnostic
configures a paired plain-H16 transform for Wgrad only; that transform does
not enter the scored Dgrad contraction and is not part of the comparison.

For each captured tensor, four stochastic candidate tensors are averaged
before computing
\begin{equation}
  E=\sum_i\left(\widehat{dX}_i-dX_i^{\mathrm{BF16}}\right)^2.
\end{equation}
The capture covers three
sequence-2,048 batches, QKV, attention-output, and FFN projections in all 32
blocks, and an independently seeded stochastic-rounding generator rather than
the training run's saved generator state. This gives 96 layer--component
comparisons per batch and 288 in total.
\localcta{} has lower error energy in all 288 comparisons. Giving each tensor
equal weight, the cross-capture geometric mean of
$E_{\mathrm{MX}}/E_{\mathrm{localCTA}}$ is 1.292.

\subsection{Excluded experiments}

The custom global \nvfp{}/\mxfp{} depth hybrid remains outside the common endpoint panels because
it diverged and retained only 16 of 32 terminal shards. The TE route with four
final BF16 blocks and the fixed-H32 operand hybrid both completed 160B-token
trajectories with resumable final checkpoints and are included in the endpoint
panels.

Diagnostic screens are also excluded from the common panels. The
low-precision output-head configurations produced no stable model-level
speedup. A \localcta{} transform screen resumed near the end of training under
a changed runtime, so it is not a causal comparison. The superseded downstream
evaluator omitted training-time RoPE scaling, while the superseded validation
stream was sensitive to the contiguous source-data region selected. Their
records are retained for audit, not used as scientific endpoints.

Run \texttt{make} in the report directory to regenerate the figures and PDF.
The checked-in data README lists each source file, generator, and verification
command; \texttt{data/SHA256SUMS} binds the publication inputs.

\subsection{Code and data availability}
\label{app:code-release}

Code, pinned kernel sources, experiment recipes, evaluation tools, and figure
builders are available at
\url{https://github.com/MrHuff/mfu-fp4-training}. Project-authored software is
released under the MIT License. The original manuscript and figures are
released under CC BY 4.0; included third-party components retain their own
terms. Training corpora, tokenizer assets, model checkpoints, and checkpoint
metadata are not redistributed. The repository instead provides hash-bound
input schemas and public recipes for binding authorized local copies without
recording credentials or private storage locations.

\end{document}